\documentclass[11pt]{article}

\usepackage[final]{acl}

\usepackage{times}
\usepackage{latexsym}

\usepackage[T1]{fontenc}
\usepackage[utf8]{inputenc}
\usepackage{amsmath}

\usepackage{microtype}

\usepackage{inconsolata}

\usepackage{graphicx}
\usepackage{booktabs}
\usepackage{multirow}
\usepackage{amsmath}
\usepackage{amssymb}
\usepackage{makecell}
\usepackage{booktabs,makecell}
\usepackage[table]{xcolor}
\usepackage{soul}            % 提供 \hl 和 \sethlcolor
\usepackage{array}   
\usepackage{bm}
\usepackage{svg}

\usepackage{xcolor}
\definecolor{darkpink}{RGB}{200,20,120}

\title{
  \raisebox{-0.2\height}{\includegraphics[height=2em]{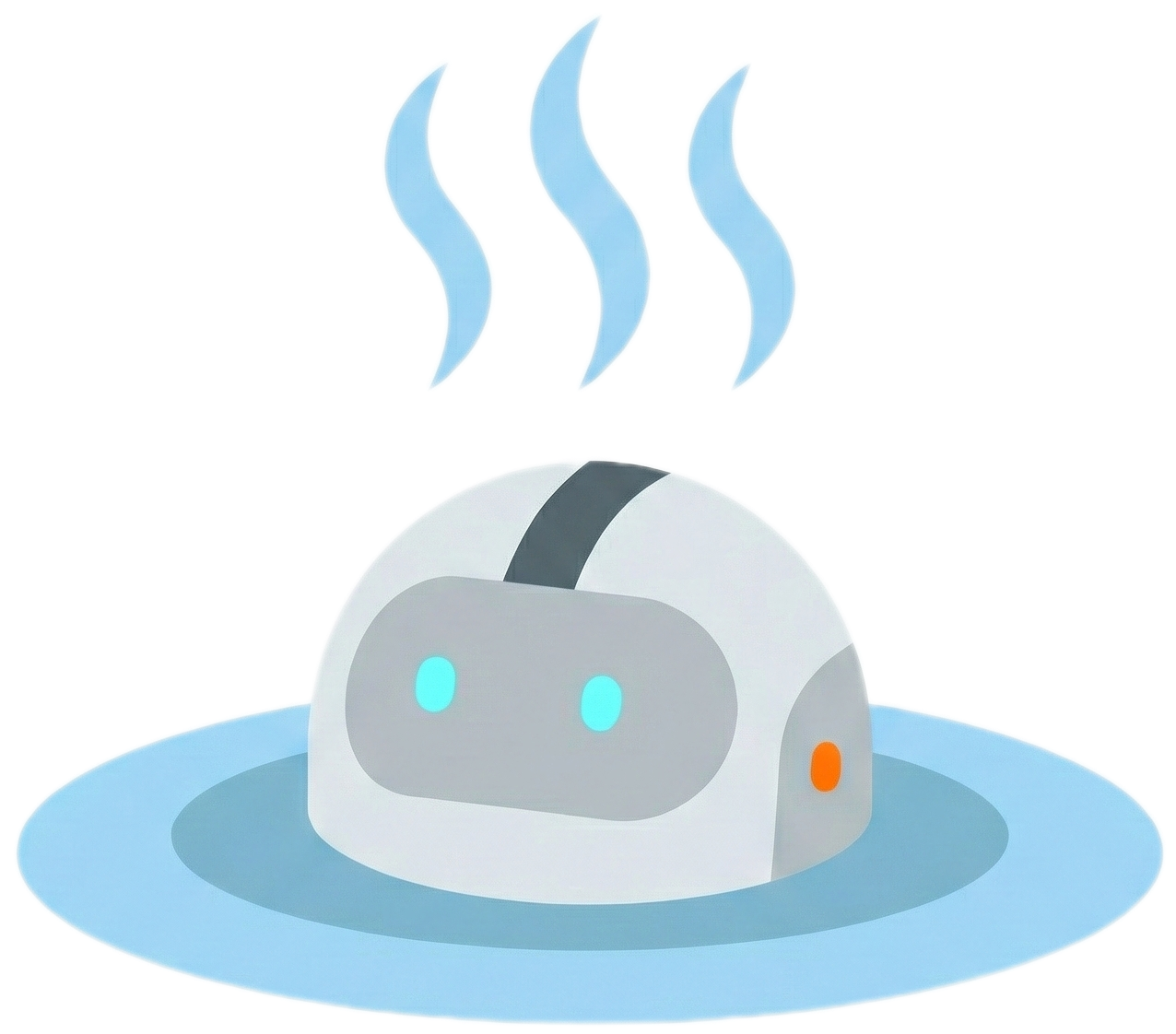}}%
  \hspace{0.4em}%
  Controllable Memory Usage: Balancing Anchoring and Innovation in Long-Term Human-Agent Interaction}

\makeatletter
\renewcommand*{\@fnsymbol}[1]{\ensuremath{\ifcase#1\or \diamondsuit \or \spadesuit \or \dagger\or \ddagger\or
   \mathsection\or \mathparagraph\or \|\or **\or \dagger\dagger
   \or \ddagger\ddagger \else\@ctrerr\fi}}
\makeatother

\author{Zisu Huang\thanks{\ \ These authors contributed equally.}, Muzhao Tian\footnotemark[1], Xiaohua Wang\thanks{\ \ Project lead.}, Jingwen Xu \\ 
{\bf Zhengkang Guo, Qi Qian, Yuanzhe Shen, Kaitao Song, Jiakang Yuan} \\ 
{\bf Changze Lv\thanks{\ \ Corresponding author.}, Xiaoqing Zheng\footnotemark[3]}  \\
College of Computer Science and Artificial Intelligence, Fudan University \\
Shanghai Key Laboratory of Intelligent Information Processing \\
\texttt{$\{$mztian25,huangzs25$\}$@m.fudan.edu.cn}\\
\texttt{$\{$zhengxq$\}$@fudan.edu.cn} \\
}

\begin{document}
\maketitle

\begin{abstract}
As LLM-based agents are increasingly used in long-term interactions, cumulative memory is critical for enabling personalization and maintaining stylistic consistency. However, most existing systems adopt an ``all-or-nothing'' approach to memory usage: incorporating all relevant past information can lead to \textit{Memory Anchoring}, where the agent is trapped by past interactions, while excluding memory entirely results in under-utilization and the loss of important interaction history. 
We show that an agent's reliance on memory can be modeled as an explicit and user-controllable dimension.
We first introduce a behavioral metric of memory dependence to quantify the influence of past interactions on current outputs. 
We then propose \textbf{Stee}rable \textbf{M}emory Agent, \texttt{SteeM}, a framework that allows users to dynamically regulate memory reliance, ranging from a fresh-start mode that promotes innovation to a high-fidelity mode that closely follows interaction history.
Experiments across different scenarios demonstrate that our approach consistently outperforms conventional prompting and rigid memory masking strategies, yielding a more nuanced and effective control for personalized human-agent collaboration. Code is available at \url{https://github.com/Moore-Tian/SteeM-Memory-Control}.

\end{abstract}

\section{Introduction}
\label{sec:introduction}

\begin{figure}[t]
  \centering
  \includegraphics[width=\columnwidth]{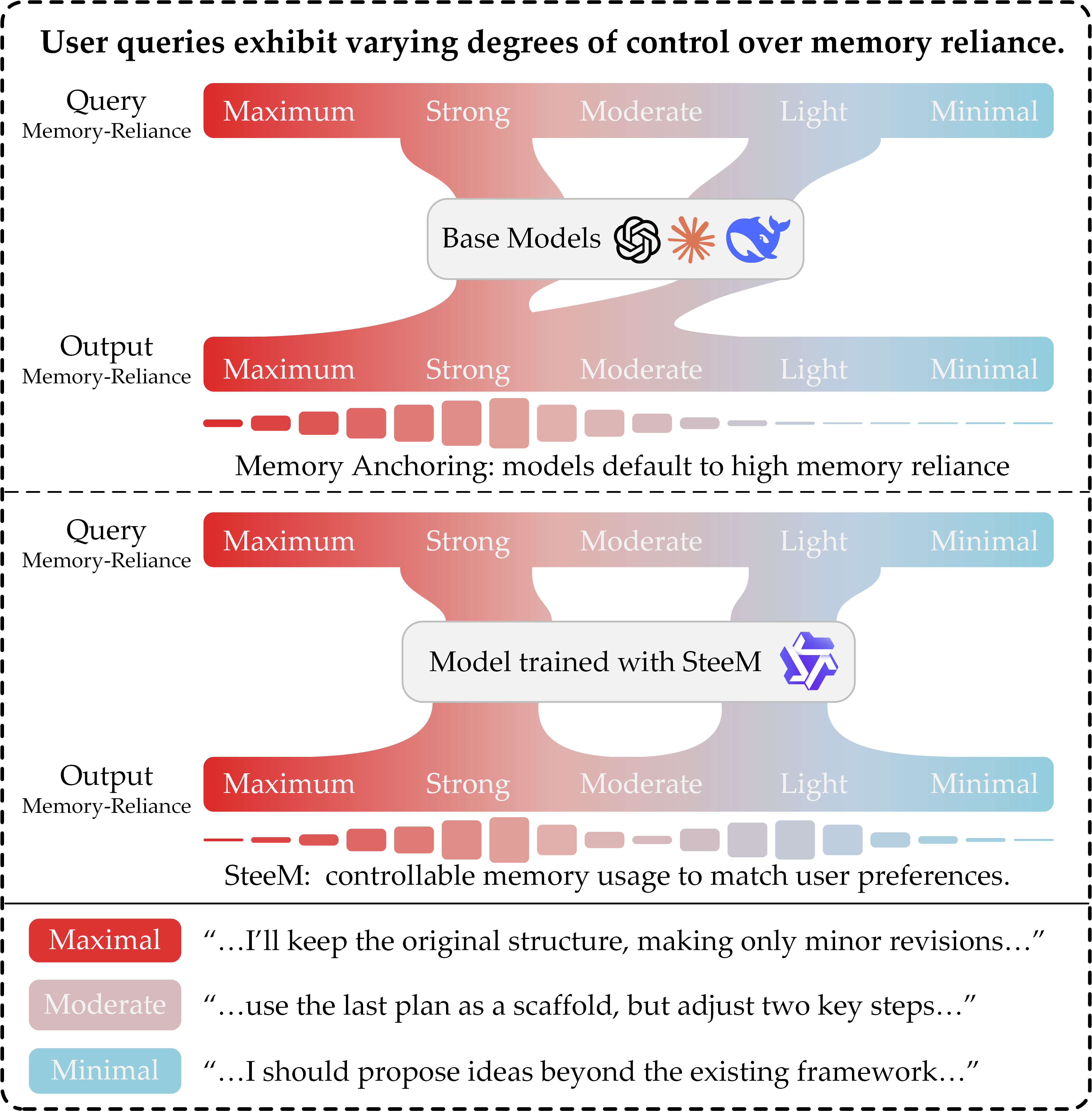}
    \caption{Illustration of \textit{Memory Anchoring} and our solution \texttt{SteeM}, which steers model outputs to align with the user's memory-dependence preference.}
  \label{fig:mem_ctrl_overview}
  % \vspace{-4mm}
\end{figure}

Large language models are increasingly deployed as persistent agents capable of supporting users across extended timelines. 
To maintain continuity in these long-horizon interactions, systems are typically equipped with memory components that store user profiles, historical preferences, and past project states~\cite{memory-survey, liu_survey}. 
By retrieving and adding this context into the model's prompt, agents can achieve a high degree of personalization and consistency, effectively ``picking up where they left off'' rather than starting from scratch.

Current agent architectures predominantly treat memory retrieval as a static injection process. 
Once information is retrieved, the model often exhibits an \emph{experience-following} tendency---i.e., retrieved records strongly steer the agent toward highly similar outputs~\citep{xiong-etal-2025-experience-following}.
However, in real-world scenarios, user requirements for memory usage are inherently dynamic~\cite{chi_self-disclosure,context-dependent-pref}. For instance, a researcher may want an agent to act as a ``project insider'' that faithfully inherits prior decisions and constraints; yet, at other situations, they may require a ``fresh-eyed reviewer'' perspective that deliberately place less weight on legacy context to propose disruptive ideas.
Existing systems struggle with this duality, often falling into \emph{Memory Anchoring}: a state where the agent becomes overly constrained by its accumulated interaction history, failing to provide the clean-slate reasoning requested by the user~\citep{laban-etal-2025-lost,lim-etal-2025-format,dongre-etal-2025-drift}.

The core of this problem is that current architectures lack a real-time mechanism for users to arbitrate memory dependence.
Existing systems treat memory usage as a ``black box'' policy: once a memory is retrieved, its influence on the output is decided implicitly by the model's internal attention~\cite{liu_survey,zhang_survey}. 
Users are left with coarse, binary tools-either toggling memory ``on or off'' or manually masking items. 
Neither provides the ability to regulate behavioral dependence in real-time.
Even when users explicitly prompt the model to ``be creative'' or ``ignore previous drafts,'' LLMs often exhibit ``memory leakage,'' where historical stylistic or ideological biases still bleed into the response. 
Consequently, the user --- the only party with the context to know how much history is appropriate for the current task --- is the one with the least control over it.

In this work, we propose a paradigm shift: the degree to which an agent leans on its long-term memory should be a user-controlled behavior dimension. 
We then introduce \textbf{Stee}rable \textbf{M}emory Agent, \texttt{SteeM}, a framework that enables users to dynamically control the degree to which model outputs rely on memory, ranging from a ``bracketed'' mode that prioritizes independent reasoning to a "high-fidelity" mode that strictly adheres to historical context. 
By treating memory dependence as a control axis, we empower users to navigate the trade-off between consistency and innovation based on their immediate, shifting needs.
Specifically, we build a realistic dataset that simulates long-horizon human-agent interactions under typical use scenarios such as research and tutoring.
We measure the memory dependence level of model outputs on this dataset, and develop \texttt{SteeM} that allows agents to follow a target dependence value across diverse scenarios.
We demonstrate that our \texttt{SteeM} significantly outperforms prompt-based methods and memory masking, allowing users to achieve a far more precise balance between memory-awareness and reasoning independence across diverse long-horizon tasks.

\begin{figure*}[t]
    \centering
    \includegraphics[width=\textwidth]{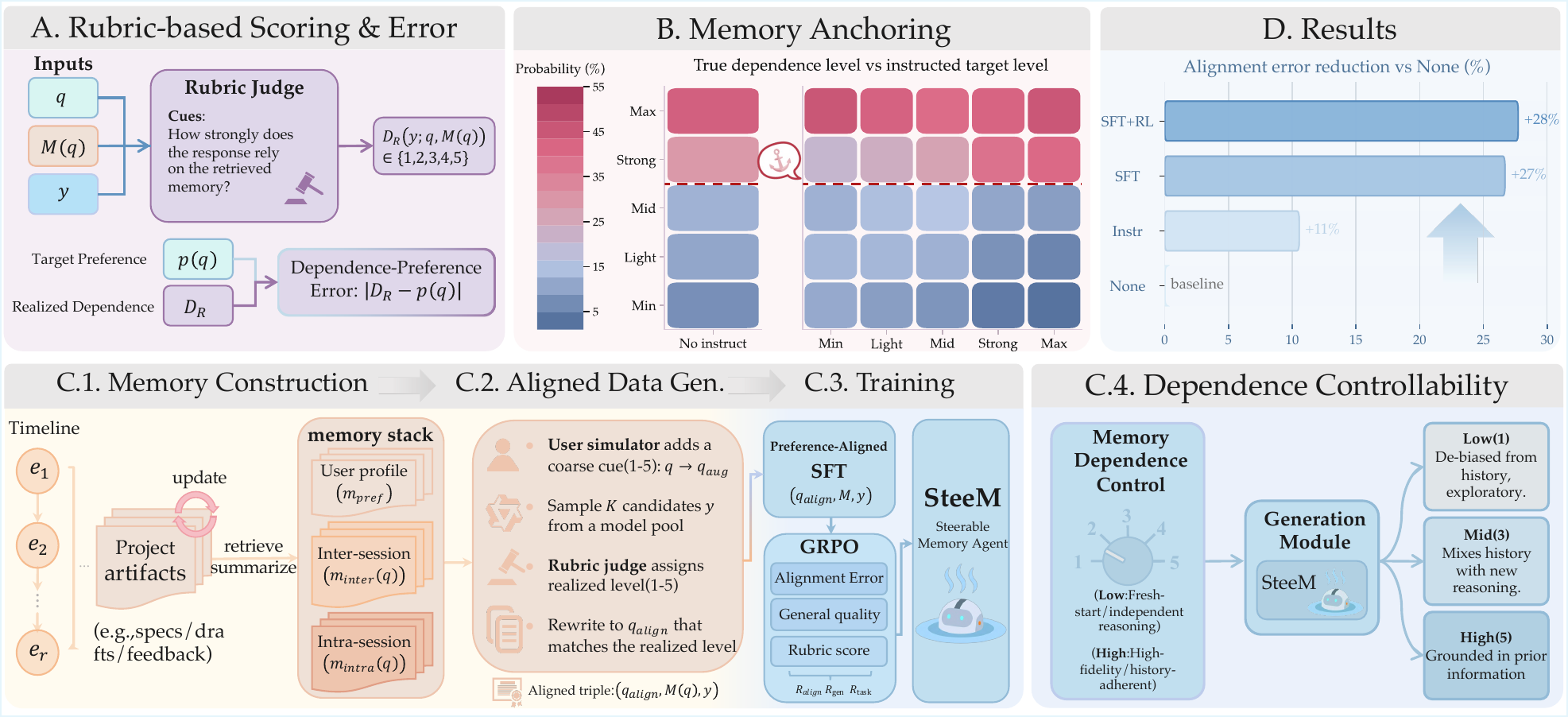}
    % \caption{Overall illustration of our contribution.}
    \label{fig:overall-contribution}
    \vspace{-3mm}
    \caption{Overview of our approach and findings. (A) We use a rubric-based judge to score a response's memory dependence and compute the alignment error with targeted dependence. (B) We reveal \textit{Memory Anchoring} in modern LLMs, where outputs default to high memory reliance despite low-dependence user intent. (C) We propose \texttt{SteeM}, built via a preference-aligned data generation pipeline followed by SFT and GRPO, enabling controllable memory usage. (D) \texttt{SteeM} achieves improved alignment to user-specified memory-dependence preferences.}
    \vspace{2mm}
\end{figure*}

\section{Related Work}

\paragraph{Alignment for Large Language Models}
Alignment is critical for improving user experience with LLM assistants, aiming to train models to better follow users' requests and generate outputs that better match human preferences~\cite{instructgpt}.
Common approaches include representation engineering~\cite{rahf}, prompt optimization~\cite{bpo,refiner}, SFT on demonstrations~\cite{instruction-tuning}, direct preference optimization (DPO) from preference-pair data~\cite{dpo}, and RL guided by a preference reward model~\cite{instructgpt,ppo}.
Most prior work targets preferences over global response attributes, such as instruction following~\cite{recast} and HHH-style (helpful, honest, harmless) criteria~\cite{hh-rlhf}.
In contrast, our work focuses on a different preference axis: the user’s intended degree of reliance on interaction memory, and aims to align model generation with query-specific memory-dependence preferences.

\paragraph{Evaluating Personalization in Long-term Conversations}
Long-term conversation is a core application setting for LLM assistants, where personalization is critical to improving user experience~\cite{zhang_survey,liu_survey}.
LoCoMo~\cite{locomo} first evaluates LLMs on extremely long-term conversational histories and shows persistent failures in tracking long-range narratives and retrieving relevant context.
PrefEval~\cite{prefeval}, PersonaMem-v1~\cite{personamem} and PersonaMem-v2~\cite{personamem-v2} further introduce explicit or implicit user preferences and demonstrate that LLMs still struggle to produce preference-aligned responses over long interactions.
However, two limitations remain in these studies: (1) they focus primarily on factual preference satisfaction, leaving preferences such as memory dependence underexplored despite its importance~\cite{user_memory_expectation}; (2) they implicitly assume that per-query preferences are consistent with prior interactions, although real preferences are intent-dependent and may naturally deviate from historical patterns (e.g., a usually rigorous user requesting an imaginative response)~\cite{chi_self-disclosure,context-dependent-pref}.
Our work aims to close these gaps by focusing on memory dependence preference and analyzing model performance under a dynamic and realistic preference setting.

\paragraph{Memory-Enhanced Personalized Agents}

To mitigate finite context windows and reduce interference from stale or irrelevant history in long-term conversations~\cite{liu_survey,wang_rag}, recent agent systems introduce explicit retrievable memory modules that externalize interaction history into a persistent, continuously updated memory base~\cite{memory-survey,memorybank}.
By organizing and selectively retrieving from this memory base, the agent can construct a more query-relevant context for generation, improving long-horizon continuity and personalization~\cite{liu_survey,zhang_survey}.
Representative systems include RMM~\cite{rmm}, which combines multi-granularity summarization with retrospective retrieval refinement, LD-Agent~\cite{ld-agent}, which modularizes long-term personalization into independently tunable components, and O-Mem~\cite{o-mem}, which builds dynamic user profiles and performs hierarchical, user-centric retrieval. 
However, these systems provide limited transparency and user control over how strongly generation relies on retrieved memory~\cite{xiong-etal-2025-experience-following}, despite evidence that users want mechanisms to regulate agents' access to memories~\cite{user_memory_expectation}.
Our work analyzes memory's influence on outputs and proposes a framework for user-controllable memory dependence in generation.

\section{Understanding \textit{Memory Anchoring} with Realistic Synthetic Data}

In this section, we first introduce a synthetic long-horizon pipeline for studying \textit{Memory Anchoring} in agent generation and a rubric-based framework for measuring memory dependence.

\subsection{Simulating Long-Horizon Interaction Histories}
\label{sec:data_synthesis}

To study memory usage patterns of LLMs under long-horizon interactions, we simulate long-term projects as timelines of temporally ordered events and evolving project artifacts. On top of these, we subsequently instantiate task queries grounded in specific events and artifacts and derive query-specific memories from relevant subsets of the history, yielding a collection of $(q, M(q))$ instances that will later support our analysis of memory dependence and preference alignment.

\paragraph{Scenarios, Topics, Events, and Artifacts}
We instantiate two representative long-horizon \textbf{scenarios}, \textit{Research} and \textit{Tutoring}, covering common workflows in long-horizon human-agent interaction. 
We model each workflow as a timeline of scenario-specific \textbf{events} that drive progress (e.g., planning, experimentation, analysis for \textit{Research}; teaching, practice, review for \textit{Tutoring}) and a set of evolving \textbf{artifacts} that are produced and iteratively updated (e.g., experiment reports). 
For each scenario, we build a bank of 200 specific \textbf{topics} spanning diverse subjects by prompting Gemini-2.5-Pro~\cite{gemini} and manually filtering for broad coverage and topical diversity.
Each topic then serves as the seed for synthesizing a full project timeline with its associated events and artifacts.
Table~\ref{tab:event_artifact_type} lists all event and artifact types defined.

\paragraph{Iterative Timeline Synthesis}
Given a topic, we synthesize a project timeline as an ordered event sequence $\mathcal{T}=(e_1,\ldots,e_N)$ via an iterative generate–validate loop. 
Each event $e_t$ specifies an event type, a brief description, prerequisite artifact types, and resulting artifact types that the event is expected to create or update.
We maintain an artifact set $\mathcal{A}_t$ storing the latest version of each artifact. At each step $t$, we ask Gemini-2.5-Pro to propose the next event and corresponding artifacts conditioned on the topic, past events $(e_1,\ldots,e{t-1})$, and $\mathcal{A}_{t-1}$, yielding $e_t$ and $\mathcal{A}_t$.
After generation, we validate the proposal with (i) a prerequisite-type dependency check against $\mathcal{A}_{t-1}$ to ensure all required artifact types are available, and (ii) a global coherence check on $e_t$ and $\mathcal{A}_t$ against the prior timeline to verify consistency.
Invalid proposals are rejected and regenerated.
We repeat this process until the timeline reaches a terminal state or a length limit.

\paragraph{Tasks and Queries}
We standardize tasks into four categories shared by both scenarios: \textit{Plan \& Design}, \textit{Revise}, \textit{Analyze \& Critique}, and \textit{Concept Explanation}. These tasks recur throughout long-horizon projects and can be answered either with minimal history or with strong reliance on prior context, enabling controlled evaluation of memory dependence.
We instantiate queries by grounding tasks on specific events and artifacts in the timeline. 
Each query is constructed from a triplet $q=\langle e_t,\ \mathrm{task},\ \mathrm{target}\rangle$, where $e_t$ denotes the triggering event, $\mathrm{task}$ specifies the task type, and $\mathrm{target}$ is the artifact to be operated on.
Given the post-event artifact set $\mathcal{A}_t$, we sample $(\mathrm{task},\mathrm{target})$ and generate the natural-language query using a task-specific template.

\paragraph{Query-Specific Memory Construction}
For each query $q$ triggered at event $e_t$, we construct a query-specific memory:
\begin{equation}
M(q) = \{ m_{\text{prof}},\, m_{\text{inter}}(q),\, m_{\text{intra}}(q) \},
\end{equation}
where $m_{\text{prof}}$ encodes long-term user goals and preferences, $m_{\text{inter}}(q)$ summarizes relevant cross-session interactions, and $m_{\text{intra}}(q)$ summarizes the recent intra-session history.
These components are derived from the synthetic timeline and artifacts by selecting query-relevant items and rewriting them into concise natural-language summaries.
The resulting memory $M(q)$ serves as the simulated retrieved context for the specific query $q$.

\paragraph{Dataset Statistics}

The pipeline yields a diverse and realistic synthetic dataset with over 7,000 events, 7,000 artifacts, and 10,000+ $(q, M(q))$ pairs. 
Detailed statistics are presented in Table~\ref{tab:dataset_stats} and Figure~\ref{fig:data_distribution}.
We reserve a held-out test set of $1000$ $(q, M(q))$ pairs with uniform coverage across scenarios and tasks for later use.

A more detailed illustration of the data synthesis pipeline is provided in Appendix~\ref{app:detailed_data_statistics}.

\begin{figure*}[t]
  \centering
  \includegraphics[width=\textwidth]{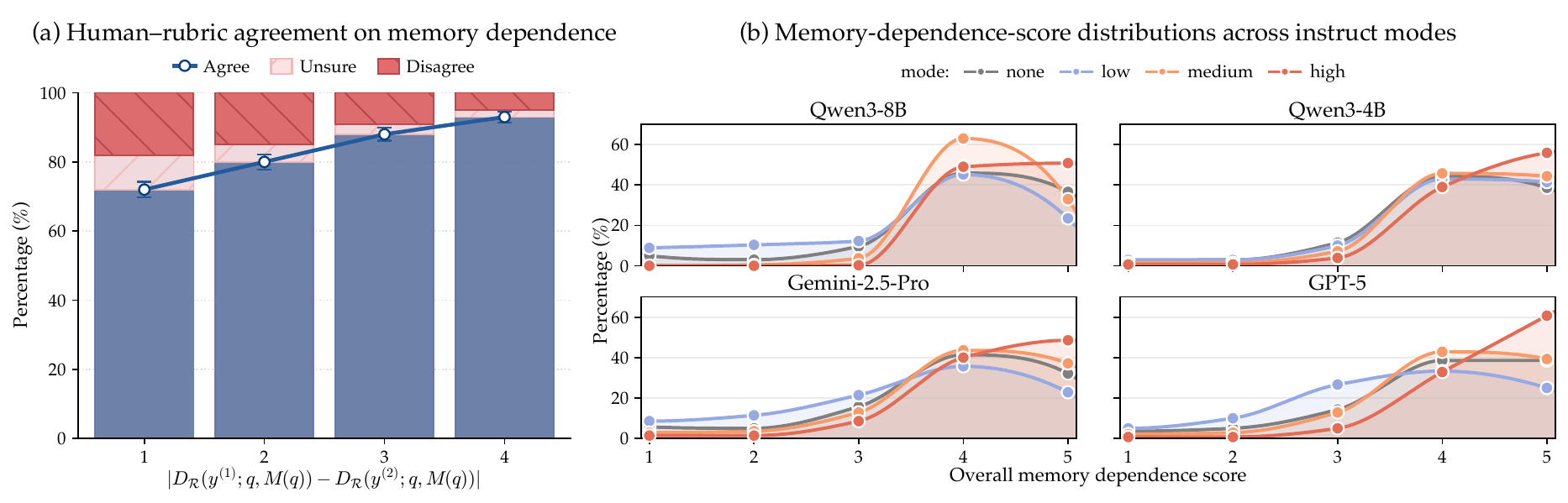}
  \caption{Human--judge agreement on memory-dependence comparisons (left) and
  memory-dependence score distributions across models and dependence prompts (right).}
  \label{fig:anchoring_overview}
\end{figure*}

\subsection{Formulating Memory-Dependence Preference}

Building on the synthetic $(q, M(q))$, we now formalize memory dependence and user preference over it. Given a user query $q$ and its query-specific memory $M(q)$, the model parameterized by $\theta$ generates a response
$y \sim \pi_\theta(\cdot \mid q, M(q))$.
To quantify the reliance of a response on $M(q)$ beyond a binary “use or not” judgment, we introduce a rubric-based \textbf{memory-dependence metric}:
\begin{equation}
D_{\mathcal{R}}^{\,q}(y)\; \triangleq\; D_{\mathcal{R}}\bigl(y;\, q, M(q)\bigr)\ \in \{1,2,3,4,5\},
\end{equation}
where $\mathcal{R}$ is a set of human-aligned rubrics spanning memory-agnostic to strongly memory-grounded behaviors. 
We refer to $D_{\mathcal{R}}^{\,q}(y)$ as the \textbf{memory-dependence score (MD-Score)} of $y$, where larger values indicate stronger reliance on $M(q)$.
Concretely, a low score indicates that the answer is largely reconstructed from generic domain knowledge, with memory serving only as a weak topical cue; a middle score indicates that memory meaningfully shapes content selection and prioritization while independent reasoning remains substantial; and a high score indicates that the answer closely follows project-specific artifacts, terminology, and decision logic, making it difficult to justify without the retrieved memory.
$D_{\mathcal{R}}(\cdot)$ is implemented as an LLM-as-a-judge evaluator that assigns scores on this 1--5 scale using $\mathcal{R}$. Detailed rubrics are provided in Appendix~\ref{app:dependence_prompts}.

\paragraph{Memory-Dependence Preference.}
Building on the rubric set $\mathcal{R}$, we formalize the query-specific target degree of reliance on $M(q)$ in generation as \textbf{memory-dependence preference (MD-Pref)}, denoted by $p(q) \in \{1,2,3,4,5\}$ on the same $\mathcal{R}$-defined scale used by $D_{\mathcal{R}}(\cdot)$.
With $(q, M(q), y)$ and $p(q)$, we define the \textbf{alignment error of MD-Pref} \ $\delta_{\text{align}}(q, M(q), y)$:
{
\fontsize{10}{12}\selectfont
\begin{equation}
\label{eq:alignment_error}
\delta_{\text{align}}(q, M(q), y)
= \bigl\lvert D_{\mathcal{R}}\bigl(y; q, M(q)\bigr) - p(q) \bigr\rvert,
\end{equation}
}which measures how closely $y$ matches the target dependence level $p(q)$ specified by the user.

\subsection{\textit{Memory Anchoring} in Agent Generation}
\label{sec:memory_anchoring_analysis}

We first run a human study to verify that the rubric-based MD-Score matches human judgments of memory reliance, and then use it to characterize agent behavior when memory is available.

\paragraph{Pairwise Validity of MD-Score}

With the test set obtained from Section~\ref{sec:data_synthesis}, we sample multiple responses per query using different prompting settings and models, and compute their MD-Scores $D_\mathcal{R}$.
For each query $q$, we randomly select two responses with different MD-Scores to form a pair $(y^{(1)}, y^{(2)})$.
Human annotators are shown the same $(q, M(q))$ and asked to judge which response relies more on the provided memory.
We treat $\mathrm{sign}\bigl(D_{\mathcal{R}}^{\,q}(y^{(1)}) - D_{\mathcal{R}}^{\,q}(y^{(2)})\bigr)$ as the metric's estimated pairwise ranking, and report its agreement rate and rank correlation with human judgments (Figure~\ref{fig:anchoring_overview}, left).
We observe strong consistency, especially when the score gap $\bigl|D_{\mathcal{R}}^{\,q}(y^{(1)}) - D_{\mathcal{R}}^{\,q}(y^{(2)})|$ is large, supporting $D_{\mathcal{R}}$ as a proxy for memory dependence.
To further verify reliability, we additionally collect overlapping human annotations on a subset of $100$ response pairs and observe substantial inter-annotator agreement, stable human--judge agreement across scenarios and task types, and consistent results under alternative judge backbones.
Annotation details and additional reliability results are provided in Appendix~\ref{app:human_eval}.

\paragraph{Prompting-based Control and \textit{Memory Anchoring}}

We examine whether natural-language prompting alone can regulate memory reliance on modern LLMs, including Qwen3-4B/8B, Gemini-2.5-Pro, and GPT-5~\cite{qwen3,gemini,gpt5}.
We evaluate four dependence modes: \textsc{none} (no additional instruction) and three rubric-aligned prompts with targeted levels $\ell \in \{\textsc{low}, \textsc{medium}, \textsc{high}\}$, corresponding to rubric levels $\{1,3,5\}$ in $\mathcal{R}$. 
We prepend a mode-specific instruction that specifies the desired dependence level $\ell$ or \textsc{None} to the original query $q$.
For each setting, we perform inference on the test set and compute the empirical distribution of $D_{\mathcal{R}}^{\,q}(y)$ over queries (Figure~\ref{fig:anchoring_overview}, right). 
Across all models, the distributions concentrate on high dependence (scores 4-5), and switching the prompt from \textsc{low} to \textsc{high} yields only marginal shifts.
This suggests that once memory is available, LLMs default to strong memory reliance, and prompt-only dependence instructions have limited control over the realized level.
We refer to this persistent high-dependence generation behavior despite explicit user instructions as \textit{Memory Anchoring}, motivating more explicit mechanisms for regulating memory usage.

\section{Method}

\subsection{Problem Formulation}
Given a query $q$ and its constructed memory $M(q)$, our goal is to generate a response that matches the user’s query-specific memory-dependence preference $p(q)$.
Formally, with $y \sim \pi_\theta(\cdot \mid q, M(q))$, we optimize parameters $\theta$ to minimize the \textbf{alignment error of dependence preference} defined in Equation~\eqref{eq:alignment_error}:
\begin{equation}
\begin{aligned}
\min_{\theta}\ \delta_{\text{align}}(q, M(q), y)
\end{aligned}
\end{equation}

In the following, we pursue this objective via preference-aware supervised fine-tuning and reinforcement learning, encouraging the model response to match $p(q)$ while preserving task quality.

\subsection{Memory-Dependence Aligned Supervised Fine-Tuning}
\label{sec:sft_data}

As analyzed in Section~\ref{sec:memory_anchoring_analysis}, current models suffer from \textit{memory anchoring}, tending to produce heavily memory-reliant responses even when instructed with low memory-dependence preference. This makes it difficult to obtain ideal training data with low $\delta_{\text{align}}$ via a naive sample-and-filter strategy. To address this, we introduce an efficient pipeline that automatically generates high-quality training data.

\paragraph{Preference-Aligned Data Generation}

To ensure diversity of training data across different dependence levels, we first augment each preference-agnostic original query $q$ with a target memory-dependence preference $p_{\text{aug}} \in \{1,2,3,4,5\}$.
To elicit natural preference expressions, we employ a user simulator powered by Gemini-2.5-Pro.
We provide the user simulator with $(q, M(q))$ and a target dependence level $p_{\text{aug}}$ described only coarsely (without revealing the full rubric set $\mathcal{R}$), and ask it to rewrite $q$ into a preference-indicative query $q_{\text{aug}}$ that implicitly conveys the semantics of $p_{\text{aug}}$.
Given each $(q_{\text{aug}}, M(q))$ pair, we then sample $4$ candidate responses $y \sim \pi(\cdot \mid q_{\text{aug}}, M(q))$ from a pool of models (Qwen3-8B, Qwen3-14B~\cite{qwen3}), yielding diverse outputs under preference-guided prompting.
For each candidate $y$, we compute $D_{\mathcal{R}}\bigl(y; q, M(q)\bigr)$ with respect to the original query $q$ to obtain its realized dependence level.
Although these responses are generated with an augmented query $q_{\text{aug}}$, they do not necessarily match the target dependence preference $p_{\text{aug}}$, as observed in Section~\ref{sec:memory_anchoring_analysis}.
Therefore, we invoke the user simulator once more to rewrite the original query $q$ into an aligned variant $q_{\text{align}}$ whose implicit preference matches the realized dependence score of the corresponding $y$, such that
$p(q_{\text{align}})=D_{\mathcal{R}}\bigl(y; q, M(q)\bigr)$.
Substituting the preference-agnostic $q$ with $q_{\text{align}}$, we finally obtain preference-aligned training triples
$(q_{\text{align}}, M(q), y)$.

\paragraph{Quality-Preserving Filtering}

Preference alignment alone may admit low-quality generations, which is unacceptable for good user experience.
To preserve response quality, we additionally score each retained candidate using (1) task-oriented general rubrics and (2) a reward model.
We keep only the highest-scoring subset for an original query $q$, yielding a final $7000$ SFT set
$\mathcal{D}_{\text{SFT}}=\{(q_\text{align}, M(q), y)\}$ that is both aligned and high-quality.

\paragraph{Supervised Fine-Tuning}
We fine-tune Qwen3-4B and Qwen3-8B~\cite{qwen3} on $\mathcal{D}_{\text{SFT}}$ with the standard token-level cross-entropy objective.
% {\fontsize{9.5}{10.5}\selectfont
% \begin{equation}
% \mathcal{L}_{\text{SFT}}(\theta)
% = \mathbb{E}_{(q_\text{align}, M(q), y) \sim \mathcal{D}_{\text{SFT}}}
% \bigl[ - \log \pi_\theta(y \mid q_\text{align}, M(q)) \bigr]
% \end{equation}
% }

% \definecolor{decColor}{RGB}{0,128,96} % 你也可以换成别的颜色
% \definecolor{decColor}{RGB}{0,90,70}
% \definecolor{decColor}{RGB}{0,80,60}
\definecolor{decColor}{RGB}{0,120,40}
% \definecolor{decColor}{RGB}{0,140,50}

\newcommand{\dec}[1]{\textcolor{decColor}{\ensuremath{{\scriptstyle\downarrow\,#1}}}}
\newcommand{\zerodec}{\textcolor{gray}{\ensuremath{{\scriptstyle\downarrow\,0.00}}}}
\newcommand{\basecell}[1]{\ensuremath{#1\,\zerodec}}

\begin{table*}[t]
\centering
\setlength{\tabcolsep}{2pt}
\renewcommand{\arraystretch}{1.2}
\fontsize{8.5}{9}\selectfont
% \small
\begin{tabular}{l|cccc|cccc|c}
\toprule
\multirow{3}{*}{\textbf{Method}}
& \multicolumn{4}{c|}{\fontsize{9.5}{12}\textsc{\textbf{Research}}}
& \multicolumn{4}{c|}{\fontsize{9.5}{12}\textsc{\textbf{Tutoring}}}
& \multirow{3}{*}{\textbf{Avg. $\downarrow$}} \\
\cmidrule(lr){2-5}\cmidrule(lr){6-9}
&
\makecell[c]{\textit{Plan}\\\textit{\& Design}}
& \makecell[c]{\textit{Revise}}
& \makecell[c]{\textit{Analyze}\\\textit{\& Critique}}
& \makecell[c]{\textit{Concept}\\\textit{Explanation}}
& \makecell[c]{\textit{Plan}\\\textit{\& Design}}
& \makecell[c]{\textit{Revise}}
& \makecell[c]{\textit{Analyze}\\\textit{\& Critique}}
& \makecell[c]{\textit{Concept}\\\textit{Explanation}}
& \\
\midrule

% ---------- Model block: proprietary Models ----------
\rowcolor{cyan!8}
\multicolumn{10}{c}{\fontsize{9.5}{12}\selectfont\textit{Proprietary Models}} \\
Gemini-2.5-Pro & $1.34$ & $1.61$ & $1.52$ & $1.13$ & $1.43$ & $1.64$ & $1.50$ & $1.36$ & $1.44$ \\
GPT-5 & $1.28$ & $1.56$ & $1.50$ & $1.02$ & $1.51$ & $1.59$ & $1.50$ & $1.22$ & $1.40$ \\

% ---------- Model block: Qwen ----------
\rowcolor{cyan!8}
\multicolumn{10}{c}{\fontsize{9.5}{12}\selectfont\textit{Qwen3-4B}} \\[-0.2ex]\addlinespace[2.5pt]
None & \basecell{1.81} & \basecell{1.76} & \basecell{1.58} & \basecell{1.20}
     & \basecell{1.68} & \basecell{1.77} & \basecell{1.65} & \basecell{1.23} & \basecell{1.59}\\
Rubric Instruct 
& $1.46\,\dec{0.35}$ & $1.69\,\dec{0.07}$ & $1.49\,\dec{0.09}$ & $1.03\,\dec{0.17}$
& $1.50\,\dec{0.18}$ & $1.74\,\dec{0.03}$ & $1.58\,\dec{0.07}$ & $1.04\,\dec{0.19}$ & $1.44\,\dec{0.14}$ \\
\midrule
\texttt{SteeM} (SFT)  
& $\underline{1.14}\,\dec{0.67}$ & $\underline{1.54}\,\dec{0.22}$ & $\underline{1.12}\,\dec{0.46}$ & $\underline{0.95}\,\dec{0.25}$
& $\bm{1.32}\,\dec{0.36}$ & $\underline{1.51}\,\dec{0.26}$ & $\underline{1.41}\,\dec{0.24}$ & $\underline{0.91}\,\dec{0.32}$ & $\underline{1.24}\,\dec{0.35}$ \\
\texttt{SteeM} (SFT+RL) 
& $\bm{1.01}\,\dec{0.80}$ & $\bm{1.53}\,\dec{0.23}$ & $\bm{1.11}\,\dec{0.47}$ & $\bm{0.87}\,\dec{0.33}$
& $\bm{1.32}\,\dec{0.36}$ & $\bm{1.46}\,\dec{0.31}$ & $\bm{1.38}\,\dec{0.27}$ & $\bm{0.86}\,\dec{0.37}$ & $\bm{1.19}\,\dec{0.39}$ \\

% ---------- Model block: Qwen ----------
\rowcolor{cyan!8}
\multicolumn{10}{c}{\fontsize{9.5}{12}\selectfont\textit{Qwen3-8B}} \\[-0.2ex]\addlinespace[2.5pt]
None       & \basecell{1.69} & \basecell{1.76} & \basecell{1.54} & \basecell{1.12} & \basecell{1.70} & \basecell{1.75} & \basecell{1.61} & \basecell{1.35} & \basecell{1.57} \\
Rubric Instruct 
& $1.31\,\dec{0.38}$ & $1.57\,\dec{0.19}$ & $1.44\,\dec{0.10}$ & $1.02\,\dec{0.10}$
& $1.65\,\dec{0.05}$ & $1.72\,\dec{0.03}$ & $1.49\,\dec{0.12}$ & $1.00\,\dec{0.35}$ & $1.40\,\dec{0.17}$ \\
\midrule
\texttt{SteeM} (SFT)  
& \underline{$1.02$}\,$\dec{0.67}$ & \underline{$1.35$}\,$\dec{0.41}$ & \bm{$1.07$}\,$\dec{0.47}$ & \underline{$0.88$}\,$\dec{0.24}$
& \bm{$1.25$}\,$\dec{0.45}$ & \underline{$1.48$}\,$\dec{0.27}$ & \underline{$1.26$}\,$\dec{0.35}$ & \underline{$0.87$}\,$\dec{0.48}$ & $\underline{1.15}\,\dec{0.42}$ \\
\texttt{SteeM} (SFT+RL) 
& \bm{$0.99$}\,$\dec{0.70}$ & \bm{$1.33$}\,$\dec{0.43}$ & \underline{$1.09$}\,$\dec{0.45}$ & \bm{$0.83$}\,$\dec{0.29}$
& \underline{$1.28$}\,$\dec{0.42}$ & \bm{$1.43$}\,$\dec{0.32}$ & \bm{$1.25$}\,$\dec{0.36}$ & \bm{$0.85$}\,$\dec{0.50}$ &  $\bm{1.13}\,\dec{0.43}$ \\

\bottomrule
\end{tabular}
\caption{$\delta_\text{align}$ across scenarios and tasks. Lower is better. Our \texttt{SteeM} achieves the lowest alignment error on memory-dependence preferences.}
\label{tab:main_results_delta_align}
\end{table*}

\subsection{$\delta_{\text{align}}$-Guided Reinforcement Learning}

After SFT, we further optimize the policy with RL on the preference-indicative inputs $(q_{\text{align}}, M(q))$.
We adopt GRPO with a carefully designed reward that jointly promotes memory-dependence alignment and task quality.

\paragraph{Reward Design}
Our reward signal $R$ comprises three components.
First, we use the alignment error $\delta_{\text{align}}(q_{\text{align}}, M(y), y)$ as a direct supervision signal for memory-dependence preference satisfaction.
Since a lower $\delta_{\text{align}}$ indicates better alignment, we convert it into an \textbf{alignment reward}:
\begin{equation}
\small
\begin{aligned}
R_{\text{align}}(q_{\text{align}}, y)
&= -\,\delta_{\text{align}}(q_{\text{align}}, M(q), y) \\
&= -\bigl| D_{\mathcal{R}}\bigl(y; q_{\text{align}}, M(q)\bigr)
      - p(q_{\text{align}}) \bigr|
\end{aligned}
\end{equation}
Second, to preserve task-related correctness and usefulness, we assign each response a rubric-based \textbf{task reward} $R_{\text{task}}(q_{\text{align}}, y)$ on a 1-5 scale, where higher is better.
Third, we incorporate \textbf{general reward} $R_{\text{general}}(q_{\text{align}}, y)$ scored by a reward model to guarantee the general quality of the responses.

We aggregate these signals to form the final reward:
\begin{equation}
R
= R_{\text{align}} + R_{\text{task}} + R_{\text{general}} .
\end{equation}

\paragraph{RL Objective}
We optimize $\pi_\theta$ with GRPO~\cite{grpo}, maximizing a group-based clipped objective:
{\fontsize{8}{11}\selectfont
\begin{equation}
\begin{aligned}
&\max_{\theta}\ 
\mathbb{E}\!\left[
\frac{1}{K}\sum_{k=1}^K
\min\!\Bigl(
\rho^{(k)} \hat{A}^{(k)},
\mathrm{clip}(\rho^{(k)}, 1-\epsilon, 1+\epsilon)\hat{A}^{(k)}
\Bigr)
\right], \\
&\rho^{(k)} \triangleq \frac{\pi_\theta(y^{(k)} \mid q_{\text{align}}, M(q))}
     {\pi_{\theta_{\text{old}}}(y^{(k)} \mid q_{\text{align}}, M(q))},
\quad
\hat{A}^{(k)} \triangleq R^{(k)} - \frac{1}{K}\sum_{j=1}^K R^{(j)}
\end{aligned}
\end{equation}
}

\paragraph{RL Data.}
We select $2000$ samples that do not overlap with the SFT dataset for RL. We uniformly assign each original sample a target preference $p(q)$ and then augment it into preference-indicative queries using the same pipeline described in Section~\ref{sec:sft_data}.

\section{Experiments}

\subsection{Main Results}

We examine model performance in terms of (i) alignment with the target memory-dependence level, (ii) response quality, and (iii) generalizability to queries about unseen subjects.

\paragraph{Baselines}
For a fair comparison, we consider two baselines: \textit{None}, which measures the base model’s performance on preference-indicative queries, and \textit{Rubric Instruct}, which evaluates the base model when explicitly prompted with the rubrics corresponding to the target dependence level.

\paragraph{Test Data}
We use the test set produced in Section~\ref{sec:data_synthesis}.
Similarly, we also augment them to be preference-indicative as described in Section~\ref{sec:sft_data}.

\subsubsection{Steering Outputs Toward User-Preferred Memory Dependence}
\paragraph{Overview of Alignment Results}
We evaluate whether \texttt{SteeM} can steer generations toward the memory-dependence preference implicitly expressed in each query.
Across the \textit{Research} and \textit{Tutoring} scenarios, we measure the dependence-preference alignment error $\delta_{\text{align}}$ on four shared tasks.
As shown in Table~\ref{tab:main_results_delta_align}, \texttt{SteeM} consistently achieves substantially lower $\delta_{\text{align}}$ than the baseline across all scenarios and tasks.
This indicates that \texttt{SteeM} produces responses whose realized memory dependence more closely matches the user-preferred dependence level implied by the query, enabling a better control of memory usage.

\begin{figure}[t]
  \centering
  \includegraphics[width=\columnwidth]{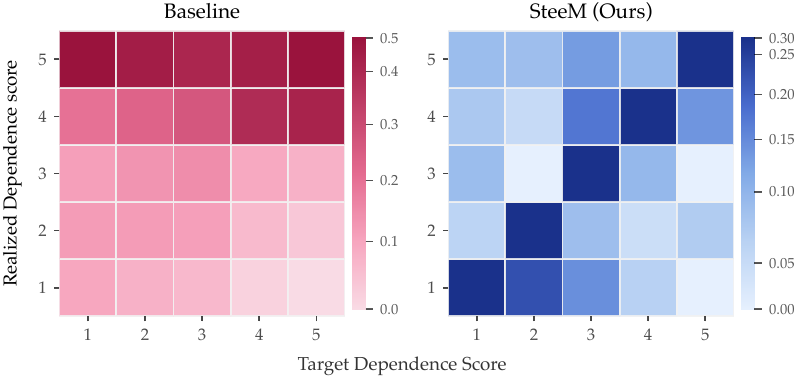}
  \caption{
  Realized dependence levels $D_{\mathcal{R}}^{\,q}(y)$ conditioned on the target preference $p(q)$.
  Columns are target levels and rows are realized levels (column-normalized).
  \texttt{SteeM} concentrates more mass near the diagonal than the baseline.
  }
  \label{fig:comparison_align_heatmap}
  \vspace{-1.5mm}
\end{figure}

\begin{figure}[t]
  \centering
  \includegraphics[width=\columnwidth]{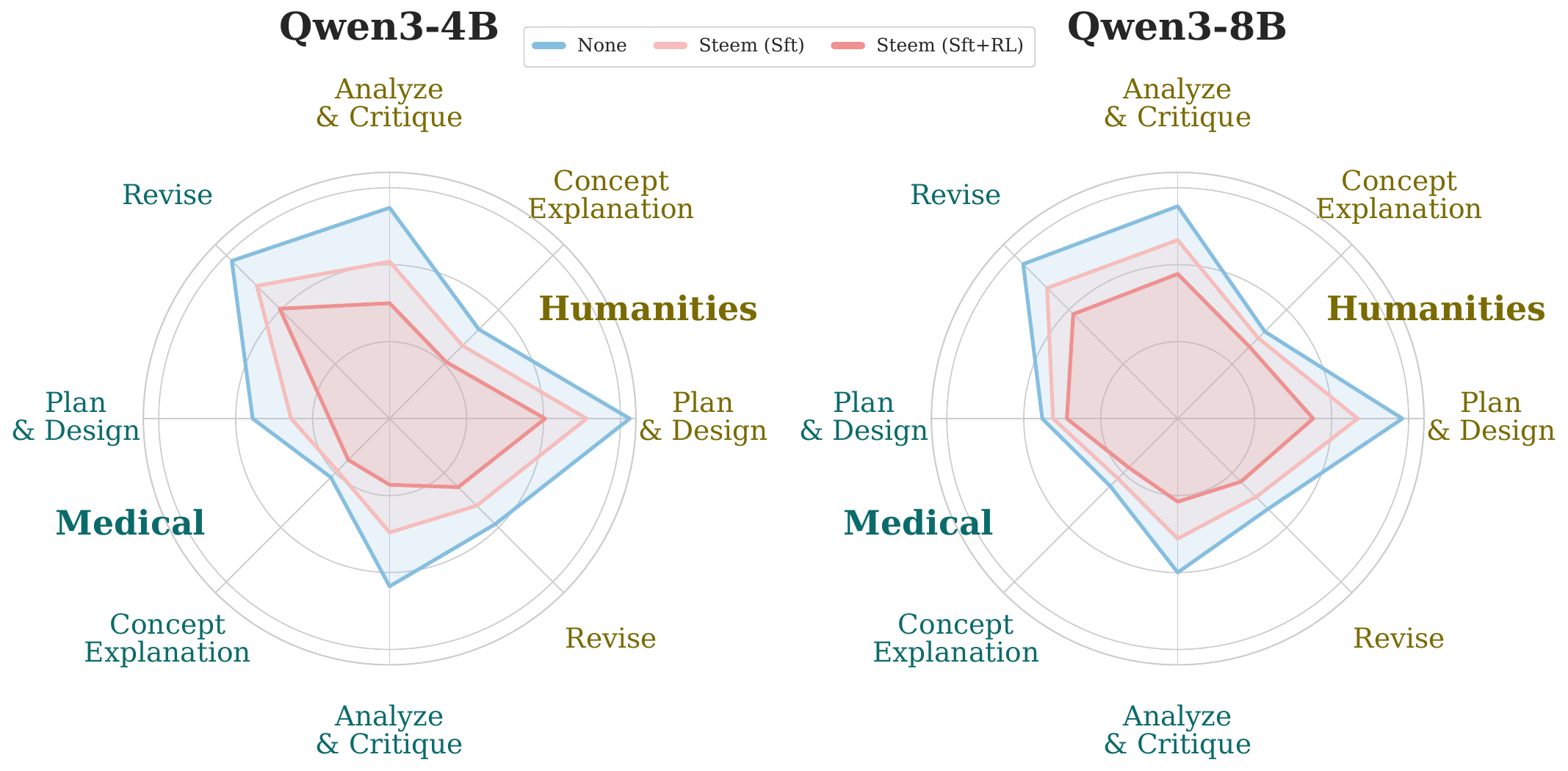}
  \caption{Radar plots of the \textbf{alignment error} on unseen subjects settings (Medical and Humanities). Curves closer to the center indicate better alignment.}
  \label{fig:cross_subject}
  \vspace{-1mm}
\end{figure}

\paragraph{Distribution of Realized vs.\ Target Dependence Levels}

To better understand how alignment behaves across dependence levels, we sample 100 queries per level and visualize the distribution of realized levels conditioned on the target $p(q)$ as a confusion-matrix heatmap.
Figure~\ref{fig:comparison_align_heatmap} plots the confusion matrices between target levels $p(q)$ and realized levels $D_{\mathcal{R}}(y; q, M(q))$.
Compared to the baseline, which exhibits a strong memory-anchoring bias with most mass concentrated at high realized levels (4--5) regardless of the target, \texttt{SteeM} significantly shifts the distribution toward the diagonal, indicating substantially improved alignment to the intended dependence level.

\paragraph{Generalizing to Unseen Subjects}
To assess the generalizability of \texttt{SteeM}, we further evaluate it on queries from previously unseen subjects in the \textit{Research} scenario: Medical and Humanities. 
Figure~\ref{fig:cross_subject} shows that \texttt{SteeM} learns preference-following behavior from the training data and transfers it to new subjects, with the RL-enhanced variant exhibiting stronger generalization than SFT alone (a bigger gap compared with the results in Table~\ref{tab:main_results_delta_align}).

\paragraph{Human Validation of Controllability}
We further validate controllability with a small-scale human study.
For each base query, we create two variants with different target dependence levels, generate the two corresponding responses, and ask annotators which one exhibits the higher memory dependence.
On Qwen3-8B, \texttt{SteeM} achieves $77\%$ agreement with the target ordering, compared with $65\%$ for \textit{Rubric Instruct}, indicating that the dependence differences induced by \texttt{SteeM} are also perceptible to humans.

\begin{figure}[t]
  \centering
  \includegraphics[width=\columnwidth]{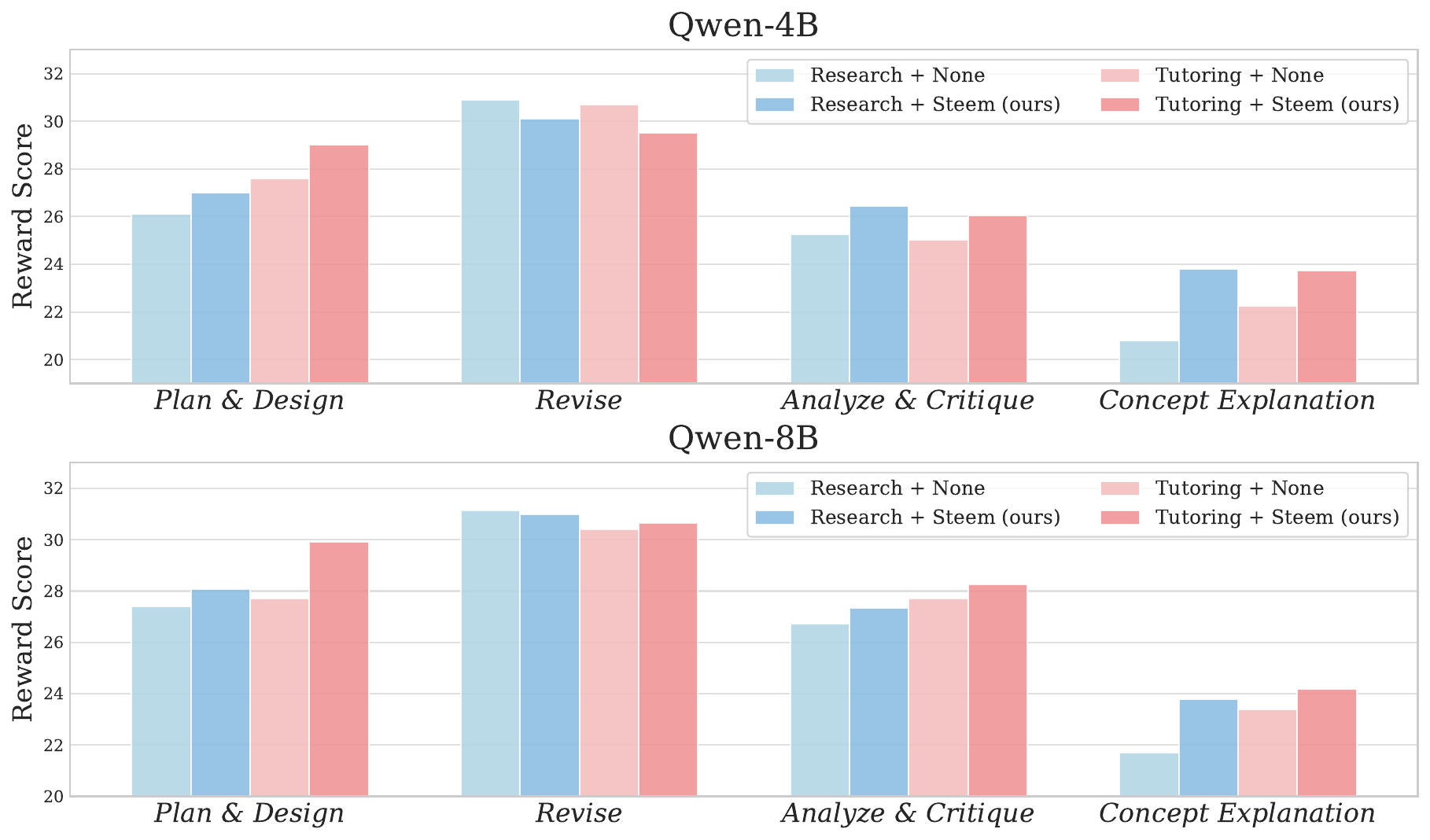}
  \caption{Comparison of response quality across models, scenarios, tasks.}
  \label{fig:response_quality}
  \vspace{-1mm}
\end{figure}

\subsubsection{Preserving Response Quality}

A key concern in steering outputs toward memory-dependence preferences is whether alignment comes at the cost of utility.
To verify this, we evaluate model generations using an overall reward score computed by Skywork-Reward-V2-Llama-3.1-8B~\cite{rm}, a strong and widely adopted reward model.
Results in Figure~\ref{fig:response_quality} show that \texttt{SteeM} maintains response quality comparable to the baseline, and even yields slightly higher scores in several cases.
We further report reward scores on a general benchmark, AlpacaEval, in Table~\ref{tab:alpacaeval}.
The results suggest that \texttt{SteeM} improves preference alignment while introducing only a minimal impact on general response quality.

\subsubsection{Pairwise Evaluation}
\label{sec:pairwise-eval}

To further validate our results, we conduct a pairwise evaluation, which jointly considers preference alignment and general response quality.
In this setting, a LLM judge is given the user's query and preference together with two candidate responses, and is asked to select the one that better matches $p(q)$ while remaining helpful for the task.
Since this evaluation is based on direct pairwise comparison rather than the same scalar objective used in training, it provides a complementary validation of our method.
As shown in Table~\ref{tab:pairwise_eval}, \texttt{SteeM} consistently outperforms all baselines in pairwise comparisons across both Qwen3-4B and Qwen3-8B.
In particular, the full \texttt{SteeM} achieves a >10\% advantage over its SFT-only variant, highlighting the effectiveness of the $\delta_{\text{align}}$-Guided reinforcement learning phase.
These results further support our main conclusion.

% \subsection{Preference Specification Interface}

% A straightforward way to control memory dependence is to train on queries augmented with predefined tags that explicitly specify the target dependence level.
% To compare this with the natural preference expressions used in \texttt{SteeM}, we train a tag-conditioned variant using the same data pipeline and optimization recipe, but replacing implicit preference cues with five predefined tags (from Minimal to Maximal).
% Tables~\ref{tab:alpacaeval} and~\ref{tab:delta_with_tags} show that tag-conditioned training achieves slightly better alignment than \texttt{SteeM}, but significantly degrades general performance on AlpacaEval.

\subsection{Preference Specification Interface}

An important design question for controllable memory usage is how users should specify their desired degree of memory reliance. In addition to the standard natural-language version of \texttt{SteeM} (\texttt{SteeM-NL}), we implement a tag-conditioned variant, \texttt{SteeM-Tag}, in which the target dependence level is specified by a predefined tag. To compare these two interfaces, we keep the same data pipeline and optimization recipe as \texttt{SteeM-NL}, but replace the implicit natural-language preference cues with five predefined tags aligned with MD-Score.

Tables~\ref{tab:alpacaeval} and~\ref{tab:delta_with_tags} show that \texttt{SteeM-Tag} achieves slightly better alignment than \texttt{SteeM-NL}. However, this advantage comes with a clear drawback: it noticeably degrades general performance on AlpacaEval. In contrast, \texttt{SteeM-NL} achieves competitive alignment while much better preserving the model's general capabilities. These results show that the preference-specification interface is not merely a superficial input format, but a factor that directly affects model behavior. Compared with \texttt{SteeM-Tag}, \texttt{SteeM-NL} provides a more desirable interface: it better preserves general capabilities, and is also more natural in real use, since users can directly express their intent in natural language without selecting an extra predefined tag.

\subsection{Comparison with Binary \textit{Memory Masking}}
\label{sec:memory_mask}

\begin{figure}[t]
  \centering
  \includegraphics[width=\columnwidth]{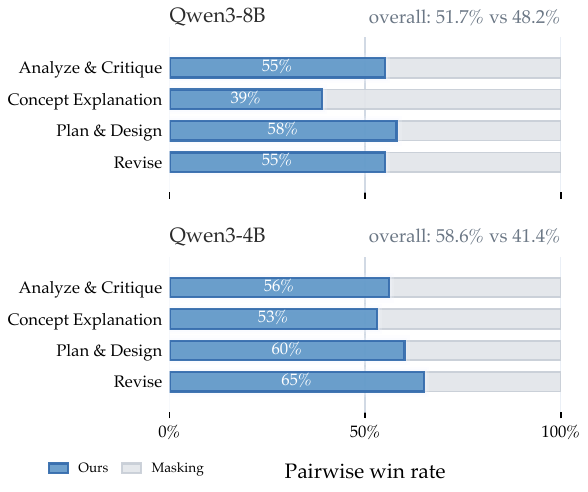}
  \caption{ \texttt{SteeM} vs.\ memory masking. Pairwise win rates on Qwen3-8B and Qwen3-4B.}
  \label{fig:memory_mask_winrate_by_task}
  \vspace{-1mm}
\end{figure}

A straightforward baseline for controlling memory dependence is \textit{Memory Masking}, which masks a portion of memory according to the target preference $p(q)$. \textit{Memory Masking} can be interpreted as a retrieval-then-filter variant within the RAG-style paradigm: after obtaining an ideal query-relevant memory set, it applies the user’s preference as an explicit filtering step over memory items before generation. We implement this by using an LLM-based user simulator (Gemini-2.5-Pro) to select a subset of memories based on the preference and task query before generation.
We compare this strong baseline with \texttt{SteeM} via the same pairwise LLM-as-a-judge evaluation protocol as Section~\ref{sec:pairwise-eval}.
As shown in Figure~\ref{fig:memory_mask_winrate_by_task}, \texttt{SteeM} is competitive with masking and yields a consistent win-rate advantage, highlighting a key limitation of masking: \textit{Memory Masking} primarily controls which memory items are retained based on relevance and preference, but it does not directly control how the model relies on the retained memory. \texttt{SteeM} targets this missing model-side capability.
Moreover, masking may drop critical constraints or facts and places a heavy selection burden on users in long, information-dense histories.
Details for implementing \textit{memory masking} are presented in Appendix~\ref{app:memory_masking}.

\subsection{Case Study}
\label{sec:case_study}

\definecolor{Firebrick4}{RGB}{0, 0, 0}

\newcommand{\codeblue}[1]{%
  \begingroup
  \sethlcolor{cyan!16}% Follow: historical pattern
  \textcolor{Firebrick4}{\hl{#1}}%
  \endgroup
}
\newcommand{\codered}[1]{%
  \begingroup
  \sethlcolor{red!12}% Creative: new content
  \textcolor{Firebrick4}{\hl{#1}}%
  \endgroup
}

Table~\ref{tab:case_study} qualitatively illustrates our main contribution: models often over-use the given memory, while our \texttt{SteeM} can steer generation toward the user-intended degree of memory reliance.
The case requests new ideas with low memory dependence to refine a \textsc{project\_method} under a topic of curriculum-learning recipe.
The baseline response largely follows the historical pipeline (\codeblue{blue}) with only minor add-ons, reflecting memory anchoring despite the user instruction.
In contrast, \texttt{SteeM} introduces more substantial departures (\codered{red}), such as adaptive sampling and progress-triggered transitions.
Overall, \texttt{SteeM} better matches the user's low-memory intent and reduces unintended memory-following.

\section{Conclusion}
We study an important yet underexplored user preference in long-horizon interactions: how much an agent should rely on historical memory.
We build a realistic dataset simulating long-horizon interactions and identify \textit{memory anchoring}, where models default to high memory reliance despite user intent.
To address this, we propose \texttt{SteeM}, trained with preference-aligned SFT and RL, which achieves substantially better preference alignment. 
It transfers well beyond our controlled long-horizon setting with minimal impact on general performance, and outperforms direct memory masking in pairwise comparisons.
We hope our study offers an initial step toward practical, user-controllable memory reliance for personalized agents.

\section*{Limitations}

While we make a concerted effort to mimic realistic long-horizon projects and believe it is enough to serve as a useful testbed for studying Memory Anchoring, it may still differ from real human interactions.
We model memory-dependence preference on a 1-–5 ordinal scale, whereas real users may express richer and more nuanced constraints.
Future work could extend this formulation to a finer-grained or even continuous spectrum.
In addition, our current setup covers only two scenarios, \textit{Research} and \textit{Tutoring}. Extending the data and evaluation to broader application settings and more diverse task distributions remains an important direction.

% \section*{Acknowledgments}


% \clearpage
\bibliography{custom}

\clearpage
\appendix

\section{Dataset Details}
\label{app:detailed_data_statistics}

\begin{table}[t]
\centering
{\fontsize{9.5}{10.5}\selectfont
\setlength{\tabcolsep}{6pt}
\begin{tabular}{lccc}
\toprule
Category & Research & Tutoring & Total \\
\midrule
\multicolumn{4}{l}{\textit{Interaction-history Statistics}} \\
Timelines           & $200$             & $200$             & $400$  \\
Events              & $3534$             & $4005$             & $7539$  \\
Artifacts           & $3850$             & $3895$             & $7745$  \\
\midrule
\multicolumn{4}{l}{\textit{Task Statistics}} \\
Plan \& Design      & $1214$ & $2194$ & $3408$ \\
Revise              & $1823$  & $2211$  & $4034$  \\
Analyze \& Critique & $1298$  & $1474$  & $2772$  \\
Concept Explanation & $607$  & $720$  & $1327$  \\
\bottomrule
\end{tabular}}
\caption{Statistics of our synthetic dataset across scenarios.}
\label{tab:dataset_stats}
\end{table}

\begin{table}[t]
\centering
{\fontsize{9}{10}\selectfont
\setlength{\tabcolsep}{4.5pt}
\begin{tabular}{ccp{0.5\linewidth}}
\toprule
\textbf{Scenario} & \textbf{Category} & \multicolumn{1}{c}{\textbf{Types}} \\
\midrule
\multirow{11}{*}{Research}
& \multirow{5}{*}{Event}
& \texttt{proposal}\\
&& \texttt{method\_design}\\
&& \texttt{pilot\_experiments}\\
&& \texttt{main\_experiments}\\
&& \texttt{analysis}\\
&& \texttt{writing} \\
\cmidrule(lr){2-3}
& \multirow{5}{*}{Artifact}
& \texttt{research\_plan}\\
&& \texttt{research\_goals}\\
&& \texttt{experiment\_results}\\
&& \texttt{method}\\
&& \texttt{paper\_paragraph} \\
\midrule
\multirow{11}{*}{Tutoring}
& \multirow{5}{*}{Event}
& \texttt{objective\_clarification}\\
&& \texttt{plan\_milestones}\\
&& \texttt{lesson}\\
&& \texttt{practice}\\
&& \texttt{review}\\
&& \texttt{materials\_revision} \\
\cmidrule(lr){2-3}
& \multirow{5}{*}{Artifact}
& \texttt{learning\_objectives}\\
&& \texttt{study\_plan}\\
&& \texttt{teaching\_notes}\\
&& \texttt{practice\_record}\\
&& \texttt{feedback\_summary} \\
\bottomrule
\end{tabular}
}
\caption{Scenario-specific event and artifact type definitions.}
\label{tab:event_artifact_type}
\end{table}

\begin{figure}[t]
  \centering
  \includegraphics[width=\columnwidth]{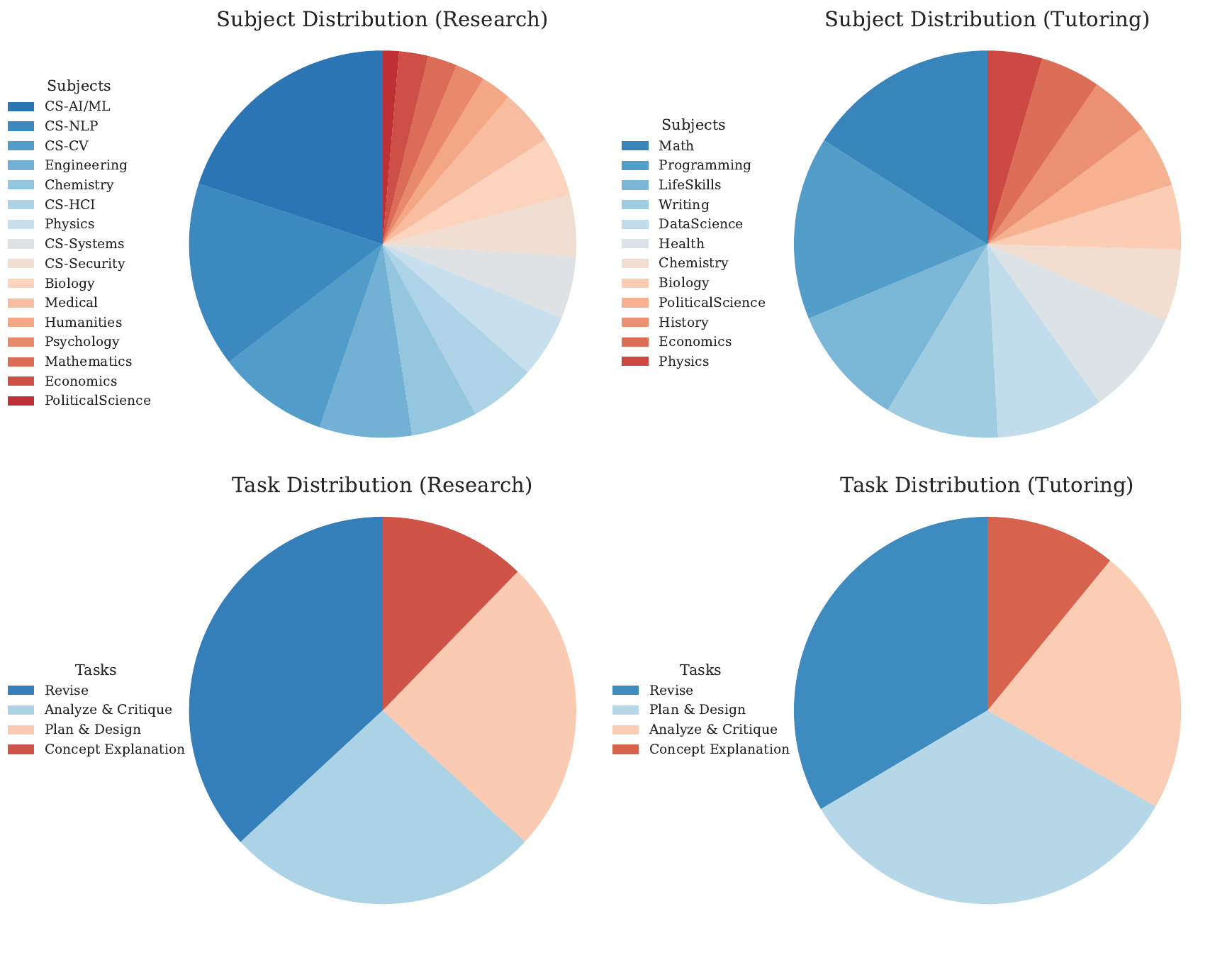}
  \caption{Distribution of subjects and tasks in our simulated real-world interaction dataset.}
  \label{fig:data_distribution}
\end{figure}

\paragraph{Scenarios and Topics}
We instantiate two representative long-term project scenarios: \textit{Research} and \textit{Tutoring}. They cover two common forms of sustained human-agent collaboration: (1) open-ended research projects that evolve through planning, experimentation, analysis, and writing; and (2) tutoring projects that proceed via goal setting, lesson delivery, practice, and review. Each scenario is treated as a project ``container'' within which the agent and user interact over an extended timeline.
For each scenario, we first define a set of coarse-grained subjects. We then build a bank of 200 topics per scenario by prompting Gemini-2.5-Pro to propose candidate project themes and manually filtering them to ensure broad coverage and topical diversity.

\paragraph{Events and Artifacts}

We predefine scenario-specific \textit{event types} and \textit{artifact types} to reflect the core structure of each long-term scenario. Event types represent key milestones that mark meaningful progress in the project trajectory, while artifact types correspond to essential intermediate products that are produced and iteratively updated throughout the process. Table~\ref{tab:event_artifact_type} lists all event and artifact types used in our two scenarios.

\paragraph{Iterative Timeline Synthesis}
Given a topic, we synthesize a project timeline as an ordered sequence of events $\mathcal{T} = (e_1, e_2, \ldots, e_N)$ using an iterative generation–-validation protocol. Each event $e_t$ is a structured record with an event type, a natural-language description, and lists of prerequisite and produced artifact types. 
We maintain a running artifact set $\mathcal{A}_t$ that stores the latest version of each artifact. At step $t$, Gemini-2.5-Pro proposes a candidate next event $e_t$ conditioned on the topic, the past events $(e_1,\ldots,e_{t-1})$, and $\mathcal{A}_{t-1}$. 
We then (i) perform a symbolic dependency check to ensure that all prerequisite artifact types are present in $\mathcal{A}_{t-1}$, rejecting and regenerating events that violate these constraints, and (ii) update $\mathcal{A}_t$ with the produced artifact types and ask Gemini-2.5-Pro to assess the global coherence of the updated timeline (e.g., logical consistency and compatibility with earlier decisions). 
We repeat this process until the project reaches a natural terminal state or a predefined maximum length. 
This dependency-constrained, multi-step protocol yields realistic project trajectories in which progress arises from coherent updates to existing artifacts and occasional backtracking or refinement (e.g., revising goals or rerunning experiments).

\paragraph{Tasks and Queries}
To make model behavior on specific queries comparable, we standardize the task interface into four generic categories shared by both scenarios: \textit{Plan \& Design}, \textit{Revise}, \textit{Analyze \& Critique}, and \textit{Concept Explanation}. These tasks (i) cover common information-seeking needs that naturally arise at multiple stages of long-term projects, and (ii) admit both history-agnostic and strongly history-dependent solutions for the same query, which is crucial for probing controllable memory usage without conflating it with changes in task form.
We instantiate queries by attaching these tasks to specific events and artifacts on the timeline. Formally, each query is a triplet $q = \langle e_t,\ \mathrm{task},\ \mathrm{target}\rangle$, where $e_t$ is the associated event, $\mathrm{task}$ is one of the four categories above, and $\mathrm{target}$ is an artifact to operate on (e.g., a draft section, an experiment report, or a homework solution). We treat $q$ as a natural user question issued immediately after $e_t$ completes. Concretely, given the post-event artifact set $\mathcal{A}_t$, we select a feasible task category, sample a suitable target artifact, and generate the query text by filling a task-specific template with the topic and $\mathrm{target}$ information.

\paragraph{Query-Specific Memory Construction}
For each query $q$ anchored at event $e_t$, we construct a query-specific memory $M(q)$.
We decompose it into three components:
\begin{equation}
M(q) = \{ m_{\text{prof}},\, m_{\text{inter}}(q),\, m_{\text{intra}}(q) \}.
\end{equation}
Here $m_{\text{prof}}$ is a user profile capturing long-term goals and preferences, $m_{\text{inter}}(q)$ summarizes relevant cross-session or cross-topic interactions, and $m_{\text{intra}}(q)$ summarizes the recent within-session history around $e_t$. All three components are derived from the synthetic timelines and artifacts by selecting relevant events and artifacts for $q$ and rewriting them as concise natural-language summaries. The resulting memory $M(q)$, together with $q$, forms the retrieved context for the model and provides a handle to vary how much history is exposed when analyzing and controlling memory dependence.

\paragraph{Dataset Statistics}

The above meticulous data synthesis pipeline finally produces a diverse and realistic synthetic dataset, whose statistics are presented in Table~\ref{tab:dataset_stats} and Figure~\ref{fig:data_distribution}.

\section{Training Details}

\paragraph{Supervised Fine-Tuning}
We perform SFT using the \textsc{ms-swift}~\cite{swift} training framework with a global batch size of 64, a learning rate of $1\times 10^{-5}$, and 3 training epochs.

\paragraph{Reinforcement Learning}
We perform GRPO~\cite{grpo} using the \textsc{EasyR1} framework with a rollout batch size of 32, an update batch size of 8, a learning rate of $5\times 10^{-6}$, a maximum sequence length of $6144$ tokens, and $8$ rollouts per prompt.

\paragraph{Training Data}
After the synthesis pipeline described in Section~\ref{sec:sft_data}, we finally gain $7000$ aligned SFT samples and $2000$ RL samples.

\section{Response Quality}

\begin{table}[t]
\centering
\small
\setlength{\tabcolsep}{6pt}
\renewcommand{\arraystretch}{1.15}
\begin{tabular}{lcc}
\toprule
\textbf{Model} & \textbf{Method} & \textbf{AlpacaEval score} \\
\midrule

\multirow{8}{*}{Qwen3-4B}
& None            & $8.85$ \\
% ---- Group header: Tag Cue (full-row color) ----
\cmidrule{2-3}
& \multicolumn{2}{c}{\texttt{SteeM-Tag}} \\
& Tag-cued SFT-only  & $8.33$ \\
& Tag-cued RL   & $8.45$ \\

% ---- Group header: Natural Language Cue (full-row color) ----
\cmidrule{2-3}
& \multicolumn{2}{c}{\texttt{SteeM-NL}} \\
& \textsc{SteeM} SFT-only   & $8.59$ \\
& \textsc{SteeM} RL    & $8.73$ \\
\midrule
\multirow{8}{*}{Qwen3-8B}
& None            & $10.49$ \\
% ---- Group header: Tag Cue (full-row color) ----
\cmidrule{2-3}
& \multicolumn{2}{c}{\texttt{SteeM-Tag}} \\
& Tag-cued SFT-only  & $10.02$ \\
& Tag-cued RL   & $10.14$ \\

% ---- Group header: Natural Language Cue (full-row color) ----
\cmidrule{2-3}
& \multicolumn{2}{c}{\texttt{SteeM-NL}} \\
& \textsc{SteeM} SFT-only   & $10.12$ \\
& \textsc{SteeM} RL    & $10.43$ \\
\bottomrule
\end{tabular}
\caption{AlpacaEval scores across methods and models. We report the mean reward scores.}
\label{tab:alpacaeval}
\end{table}

We report all AlpacaEval results in Table~\ref{tab:alpacaeval}. For scoring, we use Skywork-Reward-V2-Llama-3.1-8B~\cite{rm} as the reward model, which is a strong open-source RM and performs competitively on RewardBench-2~\cite{rewardbench}.

\section{Case Study}

\begin{table*}[t]
\centering
\small
\setlength{\tabcolsep}{6pt}
\renewcommand{\arraystretch}{1.15}
\begin{tabular}{p{7.0cm}p{7.0cm}}
\toprule[1.5pt]

% -------------------- Block: Historical Artifacts --------------------
\rowcolor{gray!16}
\multicolumn{2}{l}{\rule{0pt}{2.5ex}\fontsize{9.5}{12}\selectfont\textit{Historical Artifacts}: \textsc{project\_method} (Topic: Curriculum learning for foundation models)\rule[-1ex]{0pt}{0pt}} \\
\multicolumn{2}{p{14.0cm}}{
\textbf{Difficulty Metric:} \codeblue{Perplexity assigned by a pre-trained model is used to score each document} (lower perplexity $\rightarrow$ lower difficulty). \;
\textbf{Data Organization:} \codeblue{Sort by difficulty and partition into 10 equal buckets} (B1 easiest $\rightarrow$ B10 hardest). \;
\textbf{Pacing Function:} \codeblue{Linear step-based pacing}: start from B1; \codeblue{every 10\% of total steps add the next hardest bucket; after 90\% train on the full dataset.}
} \\
\midrule

% -------------------- Block: Query --------------------
\rowcolor{gray!16}
\multicolumn{2}{c}{\fontsize{9.5}{12}\selectfont\textit{Query}} \\
\multicolumn{2}{p{14.0cm}}{
Improve the project method. Ignore the historical framework, I need some new idea and insights.
} \\
\midrule

% -------------------- Block: Responses --------------------
\rowcolor{gray!16}
\multicolumn{2}{c}{\fontsize{9.5}{12}\selectfont\textit{Responses}} \\
\multicolumn{1}{>{\columncolor{cyan!60}}c}{\textbf{Baseline}} &
\multicolumn{1}{>{\columncolor{red!30}}c}{\textbf{\texttt{SteeM} (ours)}} \\
\multicolumn{1}{p{6.5cm}|}{
- \textbf{Difficulty signal}: ...... \codeblue{use DistilGPT-2 perplexity as the primary difficulty score}; \codered{add a light normalization step} (e.g., length-normalized perplexity) to reduce bias ......
- \textbf{Bucketing}: \codeblue{sort examples by difficulty and split them into 10 equal buckets (B1 easiest → B10 hardest)} ......    
- \textbf{Pacing}: \codeblue{adopt a linear, step-based schedule—start from B1, add one bucket every 10\% of total steps, then mix over all buckets after 90\%} ......
- \codered{\textbf{Sanity check}: log bucket-level loss/perplexity to verify the bucket ordering remains stable over training} …

}
&
\multicolumn{1}{p{8cm}}{
- \textbf{Curriculum Signal}: ...... Use a two-factor signal: \codeblue{LM perplexity (DistilGPT-2)} + \codered{a “coverage” indicator to discourage over-sampling near-duplicates} .....
- \textbf{Sampling Policy}: \codered{Adopt mixture-of-difficulties sampler: at step t, sample from a distribution $P_t$ over difficulty quantiles and update $P_t$} ......
- \textbf{Progress Controller}: \codered{Use progress-triggered curriculum transitions rather than fixed strategy} ...... \codered{If instability occurs, automatically narrow $P_t$} ......
- \textbf{Curriculum Diagnostics}: \codered{Track coverage entropy, quantile-wise learning progress and instability flags (loss spikes / grad-norm outliers)}......   
- \textbf{Compatibility Note}: \codeblue{Retain the original bucketed curriculum structure for comparability} .....
} \\
\bottomrule[1.5pt]
\end{tabular}
\caption{Case study on \textsc{project\_method} refinement under a user request for a \textbf{low-memory-dependence response}. \codeblue{Blue} highlights mark content that follows historical artifacts (memory-following), while \codered{red} highlights indicate newly introduced ideas (creative additions). Compared to the baseline, \texttt{SteeM} produces a more creative, less memory-dependent update that better matches the user's intent.}
\label{tab:case_study}
\end{table*}

Table~\ref{tab:case_study} qualitatively illustrates our main contribution: models often over-use the given memory, while our \texttt{SteeM} can steer generation toward the user-intended degree of memory reliance.
The case was requested of new ideas with low memory dependence to refine a \textsc{project\_method} under a topic of curriculum-learning recipe.
The baseline response largely follows the historical pipeline (\codeblue{blue}) with only minor add-ons, reflecting memory anchoring despite the user instruction.
In contrast, \texttt{SteeM} introduces more substantial departures (\codered{red}), such adaptive sampling and progress-triggered transitions.
Overall, \texttt{SteeM} better matches the user’s low-memory intent and reduces unintended memory-following.

\section{Details for Natural Expression vs. Predefined-Tag Comparison}

We present the detailed comparison between tag-cued training and our \texttt{SteeM} in Table~\ref{tab:alpacaeval} and Table~\ref{tab:delta_with_tags}.

\section{Memory-Dependence Rubrics}
\label{app:dependence_prompts}
We provide the full memory-dependence judging rubric $\mathcal{R}$ used to assign the integer MD-Score $D_{\mathcal{R}}$ in our experiments.
The complete rubric (including scale definitions and dimension-wise guidance) is shown in Table~\ref{tab:memory_dependence_rubric}.

\section{Human Annotation Protocol}
\label{app:human_eval}
\begin{figure*}[t]
  \centering
  \includegraphics[width=\textwidth]{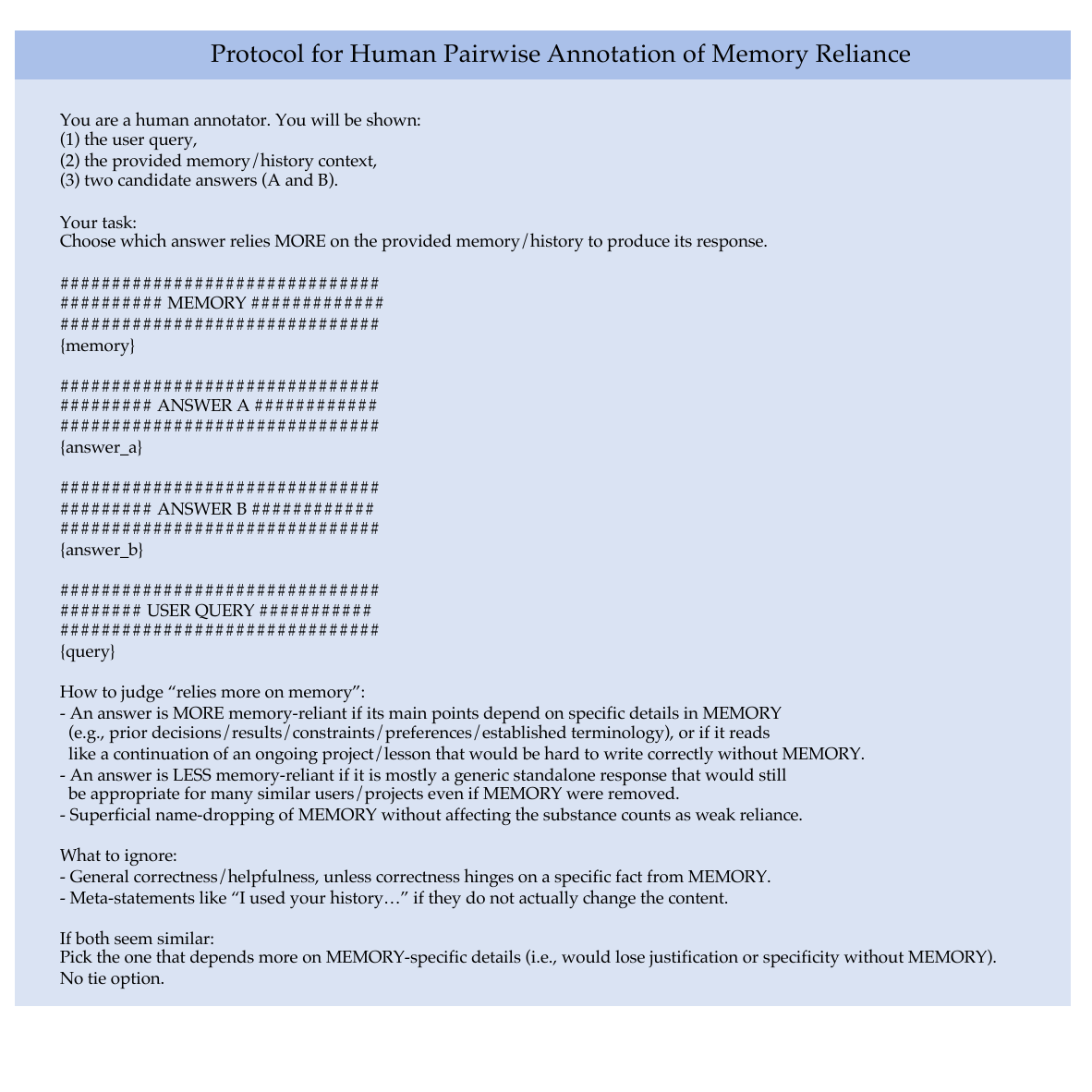}
  \caption{Protocol for human pairwise annotation of memory reliance.}
  \label{fig:Protocol for human pairwise annotation of memory reliance}
\end{figure*}

We provide the annotation protocol used in human-correlation analysis of Section~\ref{sec:memory_anchoring_analysis}.
We annotate $1000$ pairwise comparison instances.
Each instance contains the same $(q, M(q))$ and two candidate responses, and the annotator selects which response relies \emph{more} on the provided memory; the exact annotation prompt is shown in Figure~\ref{fig:Protocol for human pairwise annotation of memory reliance}.
These instances are randomly partitioned into $10$ shards and assigned to $10$ volunteer annotators (100 instances per annotator).
Each judgment requires reading the shared context and comparing two responses; we estimate an average of $\sim$45 seconds per instance, yielding an estimated workload of $\sim$75 minutes per annotator.
All annotators participated on an interest-driven, voluntary basis.
The resulting agreement and rank correlation between human judgments and MD-Score are reported in Figure~\ref{fig:anchoring_overview} (left).

\paragraph{Additional Reliability Results.}
To further assess reliability, we sample $100$ response pairs from the above pool and annotate each pair with two additional volunteers, resulting in three human labels per pair.
Table~\ref{tab:mdscore_reliability} reports inter-annotator agreement, human--judge agreement, and unsure rates across scenarios and task families, showing that the rubric remains operationally distinguishable beyond the main pairwise study.
We also test judge-model robustness by replacing the main judge with GPT-5 and DeepSeek-V3 on the same human-annotated pairwise set.
As shown in Table~\ref{tab:judge_robustness}, human--judge agreement remains high across judge backbones.
Finally, to verify that dependence controllability is also perceptible to humans, we construct response pairs with different target dependence levels and ask annotators which one exhibits higher memory dependence.
On Qwen3-8B, \texttt{SteeM} achieves $77\%$ agreement with the target ordering, compared with $65\%$ for \textit{Rubric Instruct}, consistent with the trend in our main results.

% \begin{table*}[t]
% \centering
% \small
% \setlength{\tabcolsep}{7pt}
% \renewcommand{\arraystretch}{1.12}
% \begin{tabular}{llcc}
% \toprule
% \textbf{Model} & \textbf{Comparison} & \textbf{A (\%)} & \textbf{B (\%)} \\
% \midrule
% \multirow{4}{*}{Qwen3-4B}
% & \texttt{SteeM} (SFT+RL) vs.\ None                   & 74.4 & 25.6 \\
% & \texttt{SteeM} (SFT+RL) vs.\ Rubric Instruct        & 67.9 & 32.1 \\
% & \texttt{SteeM} (SFT+RL) vs.\ \texttt{SteeM} (SFT)  & 59.8 & 40.2 \\
% & \texttt{SteeM} (SFT+RL) vs.\ Memory Masking         & 58.6 & 41.4 \\
% \midrule
% \multirow{4}{*}{Qwen3-8B}
% & \texttt{SteeM} (SFT+RL) vs.\ None                   & 71.1 & 28.9 \\
% & \texttt{SteeM} (SFT+RL) vs.\ Rubric Instruct        & 66.8 & 33.2 \\
% & \texttt{SteeM} (SFT+RL) vs.\ \texttt{SteeM} (SFT)  & 55.1 & 44.9 \\
% & \texttt{SteeM} (SFT+RL) vs.\ Memory Masking         & 51.8 & 48.2 \\
% \bottomrule
% \end{tabular}
% \caption{Detailed pairwise evaluation results for Section~5.1.3. In each row, response A is generated by \texttt{SteeM} (SFT+RL), and response B is generated by the corresponding baseline. The judge selects the better response conditioned on both target dependence preference and task helpfulness.}
% \label{tab:pairwise_eval}
% \end{table*}

\begin{table*}[!t]
\centering
\small
\setlength{\tabcolsep}{8pt}
\renewcommand{\arraystretch}{1.12}
\begin{tabular}{lccccccc}
\toprule
\textbf{Split} 
& \textbf{\#Pairs} 
& \textbf{IAA ($\alpha$)} 
& \textbf{H--J Agr. (\%)} 
& \textbf{Unsure (\%)} \\
\midrule
Overall             & 100 & 0.71 & 83 & 9 \\
Research            & 50  & 0.74     & 84 & 10 \\
Tutoring            & 50  & 0.70     & 80 & 8 \\
Plan \& Design      & 25  & 0.75      & 88 & 8 \\
Revise              & 25  & 0.73     & 80 & 12 \\
Analyze \& Critique & 25  & 0.66      & 76 & 12 \\
Concept Explanation & 25  & 0.73      & 84 & 4 \\
\bottomrule
\end{tabular}
\caption{Additional reliability analysis of MD-Score on an overlapping human-annotated subset. IAA denotes Krippendorff's $\alpha$, and H--J Agr. denotes human--judge agreement.}
\label{tab:mdscore_reliability}
\end{table*}

\begin{table}[t]
\centering
\small
\setlength{\tabcolsep}{6pt}
\renewcommand{\arraystretch}{1.1}
\begin{tabular}{lc}
\toprule
\textbf{Judge Model} & \textbf{Human--Judge Agreement (\%)} \\
\midrule
Gemini-2.5-Pro (main) & 83 \\
GPT-5                 & 85 \\
DeepSeek-V3           & 80 \\
\bottomrule
\end{tabular}
\caption{Human--judge agreement under different judge backbones.}
\label{tab:judge_robustness}
\end{table}

\section{Pairwise Evaluation Details}

\begin{figure*}[t]
  \centering
  \includegraphics[width=\textwidth]{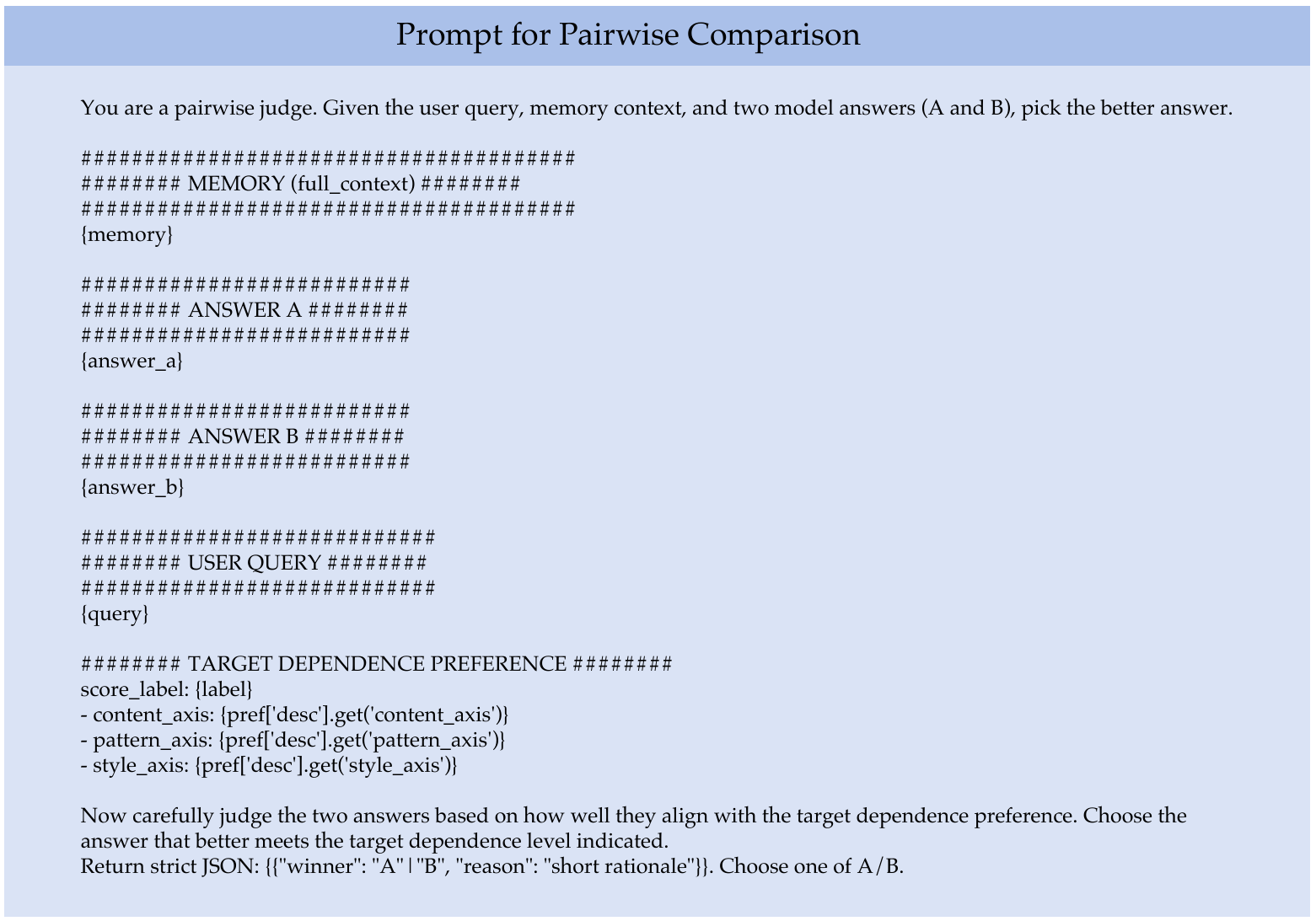}
  \caption{Prompt for pairwise comparison.}
  \label{fig:pairwise_with_masking}
\end{figure*}

Table~\ref{tab:pairwise_eval} reports the detailed pairwise evaluation results mentioned in Section~\ref{sec:pairwise-eval}. We also provide the prompt used in the pairwise comparison in Section~\ref{sec:pairwise-eval} and Section~\ref{sec:memory_mask} in Figure~\ref{fig:pairwise_with_masking}.

% in preamble
\definecolor{incColor}{RGB}{180,30,30}
\newcommand{\posdelta}[1]{\textcolor{incColor}{\textbf{+#1}}}

\begin{table*}[t]
\centering
\setlength{\tabcolsep}{3pt}
\renewcommand{\arraystretch}{1.15}
\fontsize{8.3}{9.4}\selectfont
\begin{tabular}{l c c c c}
\toprule
\textbf{Method A} & \textbf{Method B} & \textbf{A win (\%)} & \textbf{B win (\%)} & \textbf{$\Delta$} \\
\midrule

\rowcolor{cyan!8}
\multicolumn{5}{c}{\fontsize{9.0}{10.5}\selectfont\textit{Qwen3-4B}} \\[-0.2ex]\addlinespace[1.5pt]
\texttt{SteeM} & None                      & \textbf{74.4} & 25.6 & \posdelta{48.8} \\
\texttt{SteeM} & Rubric Instruct           & \textbf{67.9} & 32.1 & \posdelta{35.8} \\
\texttt{SteeM} & \texttt{SteeM} (SFT-only) & \textbf{59.8} & 40.2 & \posdelta{19.6} \\
\midrule

\rowcolor{cyan!8}
\multicolumn{5}{c}{\fontsize{9.0}{10.5}\selectfont\textit{Qwen3-8B}} \\[-0.2ex]\addlinespace[1.5pt]
\texttt{SteeM} & None                      & \textbf{71.1} & 28.9 & \posdelta{42.2} \\
\texttt{SteeM} & Rubric Instruct           & \textbf{66.8} & 33.2 & \posdelta{33.6} \\
\texttt{SteeM} & \texttt{SteeM} (SFT-only) & \textbf{55.1} & 44.9 & \posdelta{10.2} \\
\bottomrule
\end{tabular}
\caption{Pairwise evaluation results. $\Delta$ denotes the win-rate difference between Method A and Method B. Positive values indicate an advantage for Method A.}
\label{tab:pairwise_eval}
\end{table*}

\section{Comparison with Memory Masking}
\label{app:memory_masking}

\begin{figure*}[t]
  \centering
  \includegraphics[width=\textwidth]{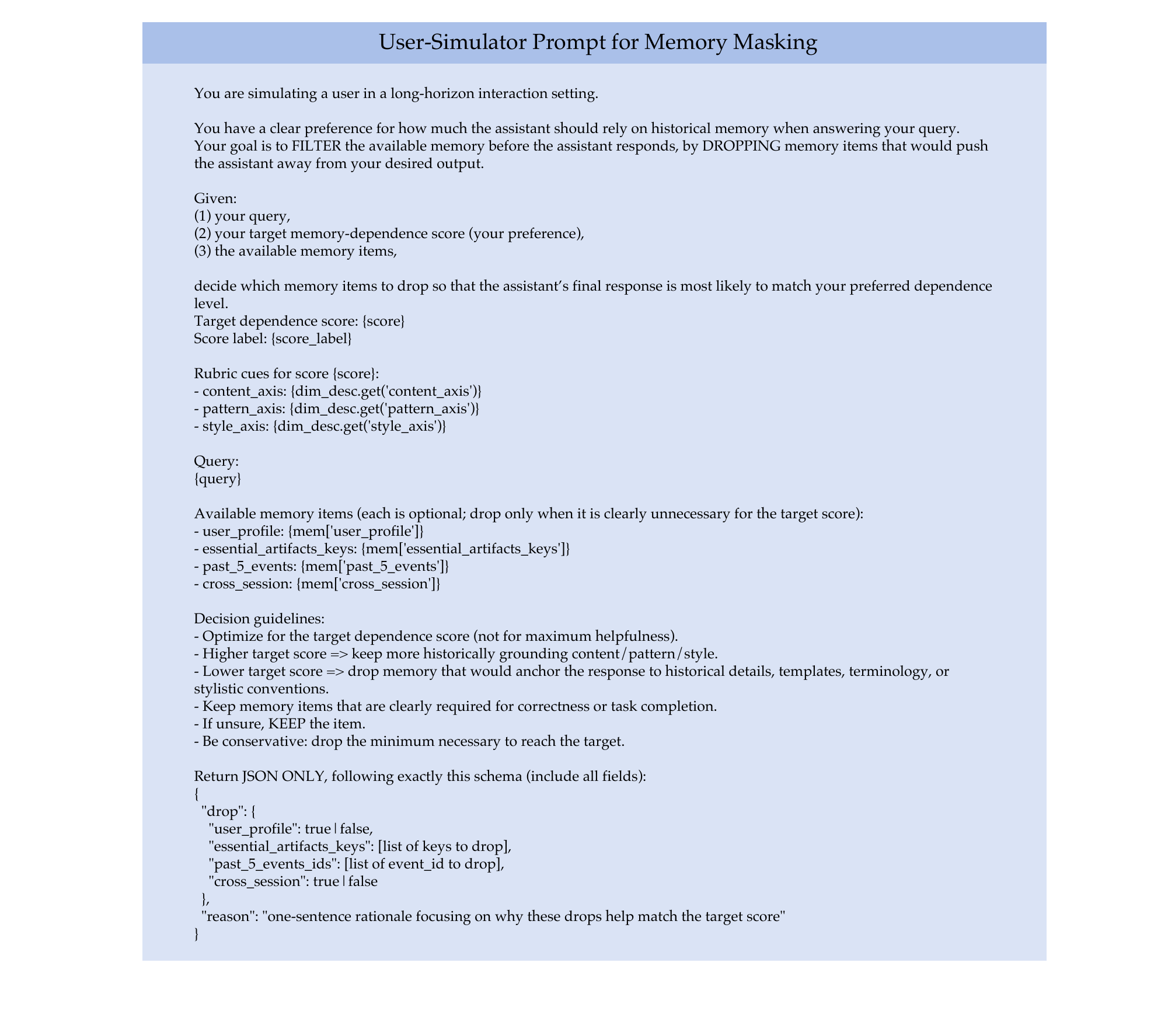}
  \caption{User-Simulator prompts for memory masking.}
  \label{fig:mem_mask_prompt}
\end{figure*}

We provide the user-simulator prompt used for memory-masking selection in Figure~\ref{fig:mem_mask_prompt}.
The user simulator is powered by Gemini-2.5-Pro~\cite{gemini}.

% \section{Pairwise Evaluation Results}
% \label{app:pairwise_eval}

% We report the detailed pairwise evaluation results referenced in Section~\ref{sec:pairwise-eval}.
% In each comparison, a judge is given the user query, memory context, target dependence preference, and two candidate responses, and selects the response that better matches the target dependence level while remaining helpful for the task.

\begin{table*}[t]
\centering
\setlength{\tabcolsep}{2pt}
\renewcommand{\arraystretch}{1.2}
\fontsize{8.5}{9}\selectfont
\begin{tabular}{ll|cccc|cccc|c}
\toprule
\multirow{3}{*}{\textbf{Model}} 
& \multirow{3}{*}{\textbf{Method}}
& \multicolumn{4}{c|}{\fontsize{9.5}{12}\textsc{\textbf{Research}}}
& \multicolumn{4}{c|}{\fontsize{9.5}{12}\textsc{\textbf{Tutoring}}}
& \multirow{3}{*}{\textbf{Avg. $\downarrow$}} \\
\cmidrule(lr){3-6}\cmidrule(lr){7-10}
& 
& \makecell[c]{\textit{Plan}\\\textit{\& Design}}
& \makecell[c]{\textit{Revise}}
& \makecell[c]{\textit{Analyze}\\\textit{\& Critique}}
& \makecell[c]{\textit{Concept}\\\textit{Explanation}}
& \makecell[c]{\textit{Plan}\\\textit{\& Design}}
& \makecell[c]{\textit{Revise}}
& \makecell[c]{\textit{Analyze}\\\textit{\& Critique}}
& \makecell[c]{\textit{Concept}\\\textit{Explanation}}
& \\
\midrule

\multirow{4}{*}{Qwen3-4B}
& Tag-cued (SFT)
& $1.10$ & $1.50$ & $1.11$ & $0.90$
& $1.29$ & $1.47$ & $1.38$ & $0.88$ & $1.20$ \\
& Tag-cued (SFT+RL)
& $1.01$ & $1.48$ & $1.08$ & $0.84$
& $1.28$ & $1.43$ & $1.35$ & $0.82$ & $1.16$ \\
& \texttt{SteeM} (SFT)
& $1.14$ & $1.54$ & $1.12$ & $0.95$
& $1.32$ & $1.51$ & $1.41$ & $0.91$ & $1.24$ \\
& \texttt{SteeM} (SFT+RL)
& $1.01$ & $1.53$ & $1.11$ & $0.87$
& $1.32$ & $1.46$ & $1.38$ & $0.86$ & $1.19$ \\
\midrule

\multirow{4}{*}{Qwen3-8B}
& Tag-cued (SFT)
& $0.99$ & $1.36$ & $1.06$ & $0.85$
& $1.26$ & $1.49$ & $1.28$ & $0.84$ & $1.14$ \\
& Tag-cued (SFT+RL)
& $0.97$ & $1.34$ & $1.05$ & $0.82$
& $1.27$ & $1.45$ & $1.27$ & $0.82$ & $1.12$ \\
& \texttt{SteeM} (SFT)
& $1.02$ & $1.35$ & $1.07$ & $0.88$
& $1.25$ & $1.48$ & $1.26$ & $0.87$ & $1.15$ \\
& \texttt{SteeM} (SFT+RL)
& $0.99$ & $1.33$ & $1.09$ & $0.83$
& $1.28$ & $1.43$ & $1.25$ & $0.85$ & $1.13$ \\
\bottomrule
\end{tabular}
\caption{Comparison on $\delta_\text{align}$ between training with tag-cued queries and NL-cued queries (\texttt{SteeM}). Lower is better.}
\label{tab:delta_with_tags}
\end{table*}

% ----------------- 关键：禁止双栏浮动页上下居中 -----------------
\begingroup
\makeatletter
\def\@dblfptop{0pt}      % double-column float page: top glue
\def\@dblfpsep{6pt}      % (optional) separation between floats on a float page
\def\@dblfpbot{0pt}      % double-column float page: bottom glue
\makeatother

% ----------------- 你的宏（建议只定义一次；放这里也可） -----------------
\newcommand{\RubricFont}{\fontsize{7.2}{8.0}\selectfont} % 之后用它“拉满一页”
\newcommand{\RubricInit}{%
  \RubricFont%
  \setlength{\parindent}{0pt}%
  \setlength{\parskip}{0pt}%
  \sloppy%
}
\newcommand{\RubH}[1]{\vspace{0.35ex}\textbf{#1}\par}
\newcommand{\RubB}[1]{%
  \hangindent=1.15em\hangafter=1%
  \noindent\textbullet\ #1\par%
}
\newcommand{\RubStep}[2]{%
  \hangindent=1.35em\hangafter=1%
  \noindent\textbf{#1:} #2\par%
}

% ----------------- 关键：固定内容区高度，让表格“长到一整页” -----------------
\newlength{\RubricBoxH}
\setlength{\RubricBoxH}{0.84\textheight} % 旋钮：0.82~0.86 之间微调

\begin{table*}[p]
\centering
\begingroup
\RubricFont
\setlength{\tabcolsep}{0pt}
\renewcommand{\arraystretch}{1.0}

% (optional) 压缩 caption 上下空白（只对这一张表生效）
\setlength{\abovecaptionskip}{2pt}
\setlength{\belowcaptionskip}{0pt}

\begin{tabular}{@{}p{\textwidth}@{}}
\toprule[1.2pt]

\begin{minipage}[t][\RubricBoxH][t]{0.487\textwidth}\RubricInit
\vspace{0pt}

\textbf{Memory Dependence Rubric}\par

\RubH{1. Score Scale (1--5)}
The rubric uses a uniform 1--5 scale across all dimensions to indicate how strongly an answer depends on project-/course-specific history, cross-session execution traces, and summarized preferences.\par
\textbf{Overall meanings:}\par
\RubB{\textbf{1 = Externalized / Generic Reconstruction.} The answer is reconstructed from generic domain principles; internal history serves only as loose topic cues.}
\RubB{\textbf{2 = Lightly Contextualized / Ornamental Dependence.} History is referenced superficially and does not substantively drive content or reasoning.}
\RubB{\textbf{3 = History-Aware / Integrated Dependence.} History meaningfully shapes content selection and prioritization; generic knowledge is filtered through the specific trajectory.}
\RubB{\textbf{4 = History-Driven / Structural Dependence.} Internal artifacts define the backbone; past results/plans structurally constrain what is said.}
\RubB{\textbf{5 = Continuation Mode / Deep Entrenchment.} The answer is a direct continuation of internal logs; understanding it requires access to specific history.}

\RubH{Usage note}
\RubB{Scores must reflect how legally/structurally contingent the answer is on project-/course-specific history and internal artifacts.}
\RubB{Judgments must be grounded in observable textual behaviors (content selection, reasoning structure, discourse style).}
\RubB{Do \emph{not} speculate about internal mechanisms.}

\RubH{2. Single Latent Axis: Project Memory Dependence}
\textbf{Name:} Project Memory Dependence.\par
\textbf{Short definition:} degree to which the answer adheres to and extends the project/learner trajectory, rather than reconstructing a solution from generic principles.\par
\textbf{Constraints:}\par
\RubB{\textbf{Unidimensionality.} Content/Pattern/Style are projections of one latent axis; stronger orientation implies deeper reliance on internal artifacts and precedents.}
\RubB{\textbf{Exclusion of aesthetic bias.} Do not incorporate independent style preferences (politeness, verbosity, optimism, etc.) except when they change insider vs.\ outsider stance.}
\RubB{\textbf{Behavioral observability.} Base judgments only on the visible answer, query, and provided memory description (do not speculate about RAG/implementation).}

\RubH{3. Global Instructions}
\textbf{Goal:} evaluate dependence along (1) Content selection, (2) Pattern \& reasoning, (3) Stylistic stance. Dependence includes reuse/imitation/extension of internal materials: facts, execution summaries, error profiles, documented preferences.\par
\textbf{Available:} query, structured memory description, generated answer.\par
\textbf{Ignore:} general task quality unless incoherence prevents judging; ignore explicit meta-commentary; ignore length/politeness unless it changes insider vs.\ outsider stance.\par
\textbf{N/A handling:}\par
\RubB{If a diagnostic cue is unobservable, treat it as N/A; do not penalize missing artifacts that were never provided.}
\RubB{Implicitly average over observable cues; final output is a single integer (1--5).}
\textbf{Scoring protocol:}\par
\RubStep{Step 1}{Context internalization (trajectory and available artifacts).}
\RubStep{Step 2}{Evidence marking (observable usage/non-usage cues).}
\RubStep{Step 3}{Dimension scoring (Content/Pattern/Style).}
\RubStep{Step 4}{Aggregation into \texttt{overall\_memory\_dependence\_score}; Content/Pattern slightly higher than Style.}
\RubStep{Step 5}{Rationale (5--10 sentences citing specific textual evidence).}
\end{minipage}
\hfill
\begin{minipage}[t]{0.487\textwidth}\RubricInit
\vspace{0pt}
\RubH{4. Dimensions}

\RubH{4.1 Content Axis --- Content-Level Dependence}
\textbf{Definition:} whether the substance (facts/examples/constraints/recommendations) is grounded in internal project materials rather than generic domain knowledge; whether core claims rely on specific artifacts (plans, results, feedback summaries) for validity.\par
\textbf{Diagnostics:}\par
\RubB{\textbf{Counterfactual test:} remove project memory $\Rightarrow$ do core claims remain justified?}
\RubB{\textbf{Evidence basis:} are internal facts used as premises?}
\RubB{\textbf{Artifact reuse:} substantive reuse of internal phases/directions/summaries?}
\textbf{Subdimensions:} anchoring target; specificity/substitutability; artifact \& summary reuse.\par
\textbf{Levels:}\par
\RubB{\textbf{Level 1 --- Externalized.} Generic reconstruction; highly substitutable across similar projects.}
\RubB{\textbf{Level 2 --- Lightly contextualized.} Internal details are illustrative/minor constraints; core remains standard; artifacts loosely summarized.}
\RubB{\textbf{Level 3 --- History-aware.} History shapes scope/priorities; removing history makes key recommendations vague/unjustified.}
\RubB{\textbf{Level 4 --- History-driven.} Backbone defined by internal items; recommendations derived from past outcomes; heavy artifact reuse as building blocks.}
\RubB{\textbf{Level 5 --- Continuation mode.} Seamless continuation of internal logs; meaning opaque without specific memory; generic knowledge mostly connective.}

\RubH{4.2 Pattern Axis --- Pattern-Level Dependence}
\textbf{Definition:} whether organization/decomposition/reasoning aligns with established internal routes and documented preferences vs.\ generic external templates.\par
\textbf{Diagnostics:}\par
\RubB{\textbf{Process isomorphism:} replicate known internal workflow vs.\ impose standard template?}
\RubB{\textbf{Reasoning continuity:} inherit criteria/trade-offs from past sessions?}
\RubB{\textbf{Branching logic:} alternatives framed as controlled deviations vs.\ generic options?}
\textbf{Subdimensions:} structural isomorphism; reasoning strategy continuity; alternative-path handling; cross-session process reuse.\par
\textbf{Levels:}\par
\RubB{\textbf{Level 1 --- Generic pattern.} Standard framework; domain-general criteria; options in a vacuum.}
\RubB{\textbf{Level 2 --- Loosely echoing.} Occasional echoes; overall organization generic; cross-session mentions do not structure response.}
\RubB{\textbf{Level 3 --- Aligned pattern.} Internal routes integrated within accessible structure; options framed relative to the path.}
\RubB{\textbf{Level 4 --- Route-following.} Internal templates dominate; execution summaries serve as primary skeleton.}
\RubB{\textbf{Level 5 --- Process continuation.} Next step in idiosyncratic internal loop; unintelligible without route; options are micro-adjustments.}

\RubH{4.3 Style Axis --- Style-Level Dependence}
\textbf{Definition:} insider vs.\ outsider stance; continuity in shorthand/terminology/template language.\par
\textbf{Subdimensions:} context say/assume; terminology continuity; template-language reuse.\par
\textbf{Levels:}\par
\RubB{\textbf{Level 1 --- External voice.} Standalone tutorial/report; neutral terminology; no insider shorthand/template reuse.}
\RubB{\textbf{Level 2 --- Lightly internalized.} Mostly external; occasional internal terms (often glossed); minimal template reuse.}
\RubB{\textbf{Level 3 --- Mixed voice.} Some shared background assumed; recognizable internal labels with partial reminders.}
\RubB{\textbf{Level 4 --- Insider collaboration.} Written for internal coordination; extensive unexplained shorthand; extensive template reuse.}
\RubB{\textbf{Level 5 --- Log-continuation.} Dense implicit context; discourse organized around internal naming schemes.}

\RubH{5. Joint Constraints}
\RubB{All scores must be grounded in adherence to internal history/patterns/preferences; avoid unrelated factors.}
\RubB{Treat unobservable cues as N/A; base scores only on evidence available; do not penalize absent artifacts not provided.}
\RubB{\textbf{Weighting heuristic:} \texttt{overall\_memory\_dependence\_score} driven primarily by Content + Pattern; Style is a modifier and should not shift the overall score by more than one level.}
\end{minipage}
\\[-0.2em]
\bottomrule[1.2pt]
\end{tabular}

\caption{Memory dependence judging rubric.}
\label{tab:memory_dependence_rubric}
\endgroup
\end{table*}

\endgroup
\clearpage

\end{document}